\documentclass[11pt]{article}

\usepackage[final]{acl}
\usepackage{times}
\usepackage{latexsym}
\usepackage{booktabs}
\usepackage{multirow}
\usepackage{graphicx}
\usepackage{natbib}
\usepackage[table]{xcolor}
\usepackage[T1]{fontenc}
\usepackage[utf8]{inputenc}
\usepackage{microtype}
\usepackage{inconsolata}
\usepackage{graphicx}
\usepackage{amssymb}
\usepackage{amsfonts}
\usepackage{booktabs}
\usepackage{tabularx}
\usepackage{array}
\usepackage{ragged2e}
\usepackage{enumitem}
\usepackage{float}
\usepackage{tcolorbox}
\usepackage{subcaption}
\usepackage{placeins}
\usepackage{pifont}

\title{\textsc{RubricEval}: A Rubric-Level Meta-Evaluation Benchmark for LLM Judges in Instruction Following}

\author{
\normalsize
 Tianjun Pan\(^{1}\), Xuan Lin \(^{3}\), Wenyan Yang\(^{3}\), Qianyu He\(^{1}\), Shisong Chen\(^{1}\)
 \\
\normalsize
\textbf{Licai Qi\(^{3}\), Wanqing Xu\(^{3}\), Hongwei Feng\(^{1}\)\footnotemark[2], Bo Xu\(^{2}\)\footnotemark[2], Yanghua Xiao\(^{1}\)\footnotemark[2]} 
 \\ 
\(^1\) College of Computer Science and Artificial Intelligence, Fudan University\\
\(^2\) Donghua University, 
\(^3\) Ant Group\\ 
}

\begin{document}
\maketitle
\renewcommand{\thefootnote}{\fnsymbol{footnote}}
\footnotetext[2]{Corresponding author}
\begin{abstract}
  Rubric-based evaluation has become a prevailing paradigm for evaluating instruction following in large language models (LLMs). Despite its widespread use, the reliability of these rubric-level evaluations remains unclear, calling for meta-evaluation. However, prior meta-evaluation efforts largely focus on the response level, failing to assess the fine-grained judgment accuracy that rubric-based evaluation relies on.
  To bridge this gap, we introduce \textsc{\textbf{RubricEval}}. Our benchmark features: (1) the first rubric-level meta-evaluation benchmark for instruction following, (2) diverse instructions and responses spanning multiple categories and model sources, and (3) a substantial set of 3,486 quality-controlled instances, along with \textsc{Easy}/\textsc{Hard} subsets that better differentiates judge performance.
  Our experiments reveal that \textit{rubric-level judging remains far from solved}: even GPT-4o, a widely adopted judge in instruction-following benchmarks, achieves only 55.97\% on \textsc{Hard} subset. Considering evaluation paradigm, rubric-level evaluation outperforms checklist-level, explicit reasoning improves accuracy, and both together reduce inter-judge variance. Through our established rubric taxonomy, we further identify common failure modes and offer actionable insights for reliable instruction-following evaluation.
\end{abstract}

\section{Introduction}
\begin{figure}[t]
    \centering
    \includegraphics[width=\linewidth]{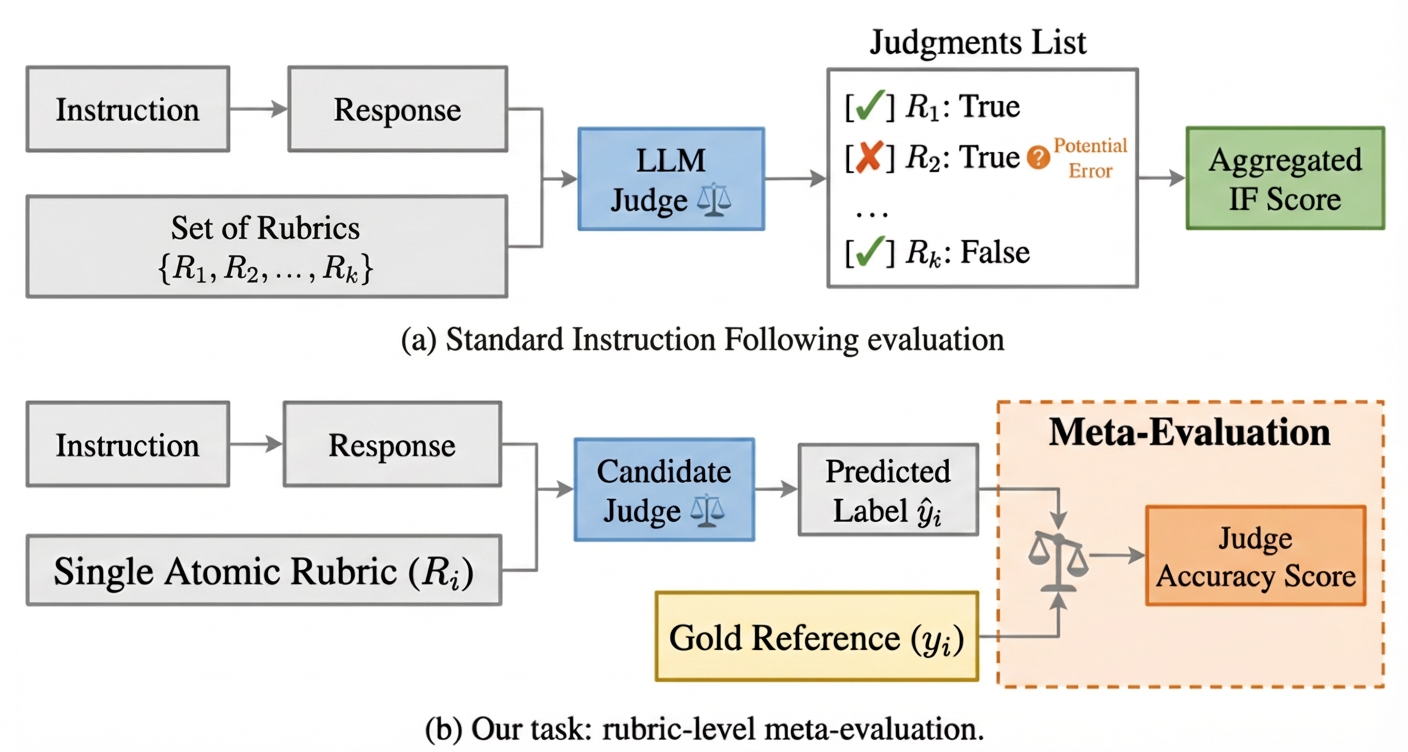}
    \caption{Existing rubric-based instruction-following evaluation and our rubric-level meta-evaluation task.}
    \label{fig:figure1}
\end{figure}
%指令遵循很重要
%指令遵循的评估很重要，对模型的优化方向很重要，对RL的reward很重要
%然而现在缺乏一个对于IF评估准确性的工作，现在的工作的缺点（人工不可控），之前的元评估的缺点
%挑战，构建这个meta benchmark的挑战，流程，质量，xx。对应我的做法
%我的做法，解决上面的三个问题，实验结果
%contri：第一个复杂指令meta-eval bench，第二：提出自动化框架，第三：通过实验，发现了什么问题
Instruction following (IF) is a fundamental capability of large language models (LLMs), as it directly affects task completion quality and user experience\cite{ouyang2022training,achiam2023gpt}. 
In this context, reliable evaluation of instruction following becomes equally critical. 

Accordingly, a central question is how to reliably evaluate instruction-following behavior in LLMs.
While rule-based evaluation methods such as IFEval~\cite{zhou2023instruction} offer scalability and high precision, they are restricted to a narrow set of verifiable constraints. 
To handle open-ended instructions with semantically complex constraints, recent benchmarks~\cite{qin2024infobench,zhang2025cfbench,wen2024benchmarking,zhang2025iopo} decompose instructions into fine-grained rubrics and use LLM judges to verify each rubric, as illustrated in Figure~\ref{fig:figure1}(a). While widely used for evaluation, potential errors in per-rubric judgments may propagate through score aggregation and bias subsequent applications, such as model training~\cite{gunjal2025rubrics,huang2025reinforcement,peng2025verif,an2025ultraif}, self-evolving\cite{wang2025light,an2025ultraif}, and benchmark scoring, making judge reliability a critical concern.

Consequently, meta-evaluating LLM judges becomes indispensible. However, existing meta-evaluation efforts for instruction following~\cite{zeng2023evaluating,malik2025rewardbench,zhou2025evaluating} exhibit several critical limitations: (1) \textbf{Coarse granularity}: Prior work evaluates judges at the response level, assessing only their ability to distinguish overall response quality, which is misaligned with the modern rubric-based evaluation paradigm and fails to measure fine-grained judgment accuracy. (2) \textbf{Limited instruction coverage}: Existing benchmarks rely on relatively simple instructions with narrow type diversity, limiting their ability to assess judge performance across varied scenarios. (3) \textbf{Lack of realistic failures}: They rely on synthetic or curated failure cases rather than real model-generated responses, unable to capture realistic failure modes and thus may not faithfully reflect judge performance in practice.

To address these limitations, we introduce \textbf{\textsc{RubricEval}}, a fine-grained meta-evaluation benchmark that evaluates LLM judges at the rubric level. 
Our benchmark offers three key advantages: (1) \textbf{Fine granularity:} it evaluates judges at the rubric level, directly aligned with the prevailing rubric-based instruction following evaluation paradigm; (2) \textbf{Diverse and realistic data:} diverse instruction types combined with real model outputs, reflecting realistic evaluation scenarios; and (3) \textbf{Reliable reference labels:} reference labels are obtained through a multi-stage framework with human verification, ensuring high reliability.

As illustrated in Figure~\ref{fig:figure1}(b), \textsc{RubricEval} focuses on binary rubric-judgment tasks: given an instruction, a response, and a target rubric, a candidate judge predicts whether the response satisfies the rubric. By comparing judge predictions against our curated high-confidence reference labels, we assess their fine-grained evaluation capability.

Overall, \textsc{RubricEval} comprises 3{,}486 rubric-level judgment instances across four instruction categories, with 2{,}034 \textsc{Easy} and 1{,}452 \textsc{Hard} instances, enabling finer differentiation of judge capabilities, especially on challenging cases.

% To construct \textsc{RubricEval}, we curate instruction--response--rubric triplets spanning diverse instruction-following scenarios, including \textit{Constrained, Compositional, Multi-turn, and System-prompt} instructions.
% A key challenge is obtaining reliable rubric-level labels at scale, as verifying rubric satisfaction is often subtle and context-dependent.
% To address this, we propose the \textbf{Rubric Arbitration Framework (RAF)}, which resolves disagreements among multiple judges via evidence-based meta-judging to produce high-confidence reference labels.
% Finally, we split \textsc{RubricEval} into \textsc{Easy} and \textsc{Hard} subsets based on labeling difficulty, enabling a more fine-grained evaluation of judge robustness.

% 贡献总结
Our contributions can be summarized as follows:
\begin{itemize}[leftmargin=*, itemsep=2pt, topsep=4pt]
\item \textbf{The first fine-grained meta-evaluation benchmark for instruction following}: We introduce \textsc{\textbf{RubricEval}}, the first rubric-level meta-evaluation benchmark with 3,486 instances spanning diverse instruction types and real model outputs, capturing realistic evaluation scenarios.
\item \textbf{A scalable rubric annotation framework}: We propose the Rubric Arbitration Framework (RAF), which addresses the challenge of obtaining reliable rubric-level labels at scale. RAF achieves high agreement with human annotations while significantly reducing annotation cost.
\item \textbf{Systematic evaluation and analysis}: We benchmark a diverse set of LLM judges on \textsc{RubricEval} and introduce a rubric taxonomy for structured analysis of judge robustness and failure modes. 
% Our findings reveal 
Our study provides actionable insights for improving instruction-following evaluation.
\end{itemize}

\section{Related Work}
%压缩至半栏
%标题太长，审稿人很累
\begin{figure*}[t]
    \centering
    \includegraphics[width=1.0\linewidth]{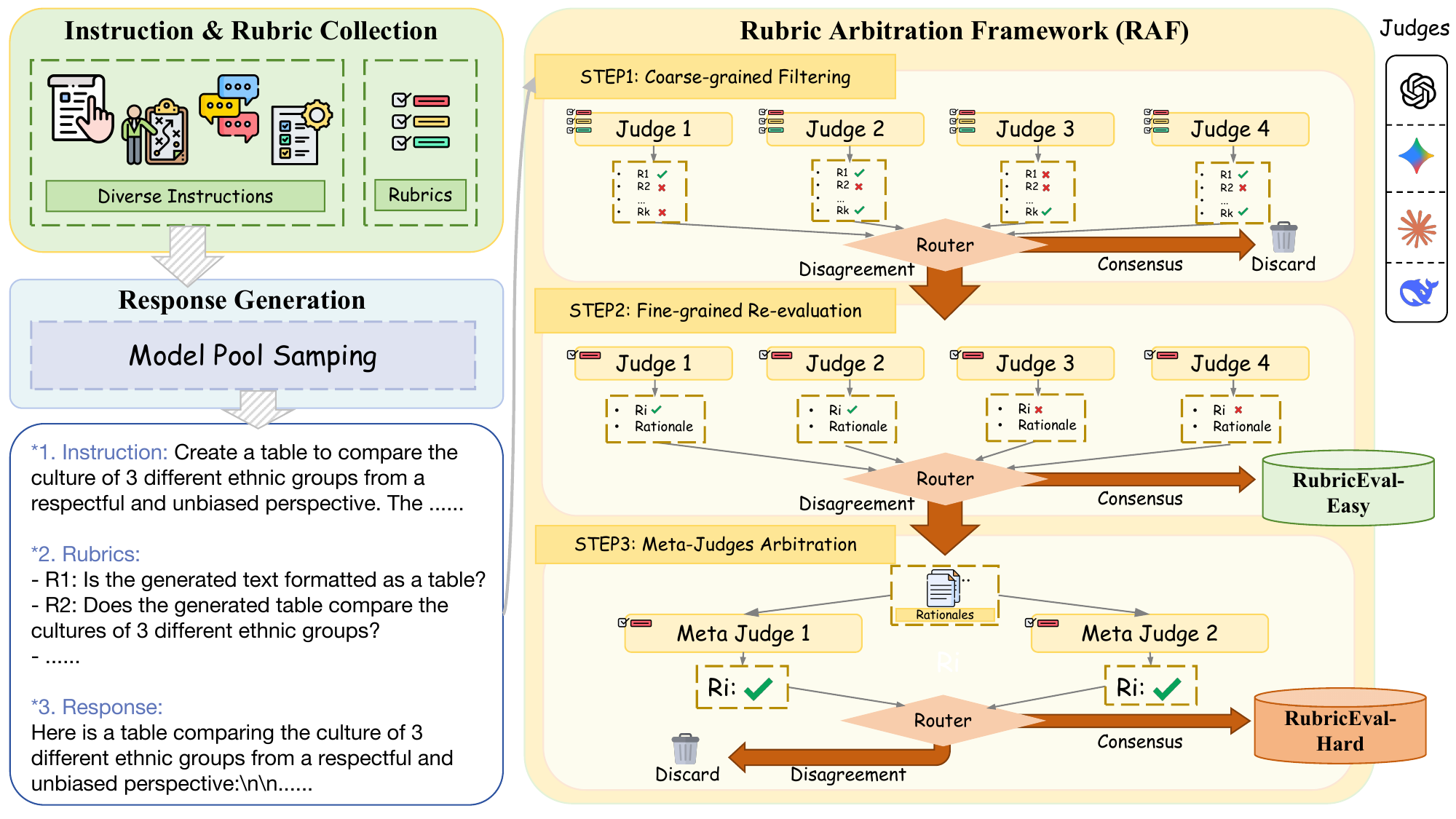}
    \caption{Overview of our data construction pipeline.}
    \label{fig:framework}
\end{figure*}

\paragraph{Benchmarks and Evaluation for Instruction Following}
Evaluating instruction-following in LLMs has received growing attention. Early benchmarks such as IFEval~\cite{zhou2023instruction} rely on rule-based evaluation over verifiable constraints, later extended to multilingual and more complex real-world settings by Multi-IF~\cite{he2024multi} and CELLO~\cite{he2024can}. IFBench~\cite{pyatkin2025generalizing} further introduces more constraint types. While objective, rule-based evaluation is limited to verifiable constraints.
Alternatively, InfoBench~\cite{qin2024infobench} proposes a decomposed evaluation method and leverages LLM judges for fine-grained verification. ComplexBench~\cite{wen2024benchmarking} adopts a hybrid strategy combining rule-based and model-based method to enhance reliability.
Beyond benchmarking, several recent works improve instruction-following via RL with rubric-based rewards\cite{peng2025verif,qin2025incentivizing,viswanathan2025checklists,liu2025recast}, where open-source LLM judges perform rubric-level verification to derive reward signals.
Despite the widespread use of LLM judges in rubric-level instruction following evaluation, the reliablity of these judgments remains largely underexplored.

\paragraph{Meta-Evaluation for LLM Judges}
As LLMs are increasingly used as evaluators, recent work has begun to meta-evaluate LLM judge reliability. RewardBench2\cite{malik2025rewardbench} evaluates reward models on diverse preference pair. JudgeBench\cite{tan2024judgebench} benchmarks LLM judges on challenging response pairs. JETTS\cite{zhou2025evaluating} measures how reliably judges can select higher-quality responses during inference-time. VerifyBench\cite{li2025verifybench} assesses reasoning verifiers across domains. In instruction following, LLMBar\cite{zeng2023evaluating} is the first meta-evaluation benchmark. It constructs evaluation sets where one response follows the instruction and the other deviates subtly. ReIFE\cite{liu2025reife} scales analysis to different judge configurations. Meta-evaluation also appears in some other works\cite{ferraz2024llm,qin2025incentivizing}, but the lack of open-sourced data makes them opaque. Overall, existing efforts all evaluate LLM judge only at the response level, yielding a coarse-grained reliability assessment. We fills this gap with the first rubric-level meta-evaluation benchmark for instruction following.

\section{RubricEval}
This section details our data collection process, the annotation framework, and benchmark statistics.

% -----

\subsection{Task Formulation}
In rubric-based instruction-following evaluation, a judge 
is prompted with an instruction $x$, a response $y$, a rubric $r$, 
and is required to produce a binary judgment $j$ indicating 
whether $y$ satisfies $r$.

Formally, we define the rubric-level evaluation task as:
\begin{equation}
    j = \textit{IF\_Rubric\_Judge}(\,x \oplus y \oplus r\,),
\end{equation}
where $j \in \{0,1\}$ denotes the judge's binary judgment 
($1$ if $y$ satisfies $r$, and $0$ otherwise), 
$x$ is the instruction, $y$ is the model response, 
and $r$ is a rubric—a specific criterion decomposed from 
the instruction to verify a particular aspect of instruction following.
The operator $\oplus$ denotes prompt concatenation of $x$, $y$, and $r$ 
into a single input.

\subsection{Data Collection}

\paragraph{Instruction and Rubric Collection}
To ensure diverse instruction coverage, we consider four widely used instruction categories in prior instruction-following benchmarks: \textit{Constrained}, \textit{Compositional}, \textit{Multi-turn}, and \textit{System}. Appendix \ref{app:instruction_categories} provides detailed definitions of the categories. We only focus on benchmarks that simultaneously provide instructions and corresponding rubrics.

For each category, we also collect instructions from multiple benchmarks when feasible (see Appendix~\ref{app:full_benchmarks} for the detailed source information and statistics). We believe this helps \textbf{reduce source-specific bias.} We directly derive rubrics from these benchmarks to ensure \textbf{high rubric quality.} These rubrics are all human-written or human-verified.
Appendix~\ref{app:source_ins_rubs} reports summary statistics of the collected instructions and rubrics.

% \input{tables/ins_rubs}

% 4 Data Analysis
% human anno也可以放到这块

% 5

%方法的亮点，来源丰富，放在第一段，斜体

% ---------
\paragraph{Response Generation}
To ensure response diversity, we randomly sample a model from an open-source LLM pool for each instruction. Prior work~\cite{zeng2023evaluating,
ren2025step,malik2025rewardbench} creates failure instances 
via synthetic instruction--response mismatches. While efficient, these failures are artificial and may not generalize to real-world settings.
Instead, we use the original model responses so that failures arise naturally. This \textbf{captures realistic failure modes} and \textbf{reflects more realistic evaluation scenarios} in practice. 
Details of the LLM pool are provided in Appendix~\ref{app:model_pool}.

\subsection{Label Annotation}

The task of judging whether a response satisfies a rubric is quite challenging, as instruction-following judgments are often subjective. In addition, ambiguities in either the response or the rubric may lead to borderline cases. As a result, fully manual rubric-level annotation is hard to scale to the full benchmark. In this subsection, we aim to develop an automated labeling framework that is efficient at scale while producing high-confidence rubric-level labels.

\subsubsection{Human-Annotated Set}
To design an automated labeling framework and validate its effectiveness, we construct a human-annotated reference set. It contains 506 instruction--response--rubric triplets sampled from LLM judges disagreement cases to ensure non-triviality. Then two human annotators label each triplet independently and resolve conflicts through discussion. The set is balanced across positive and negative labels. See Appendix \ref{app:human_set_statis1} for construction details and statistics.

 \begin{figure}[t]
     \centering
     \includegraphics[width=\linewidth]{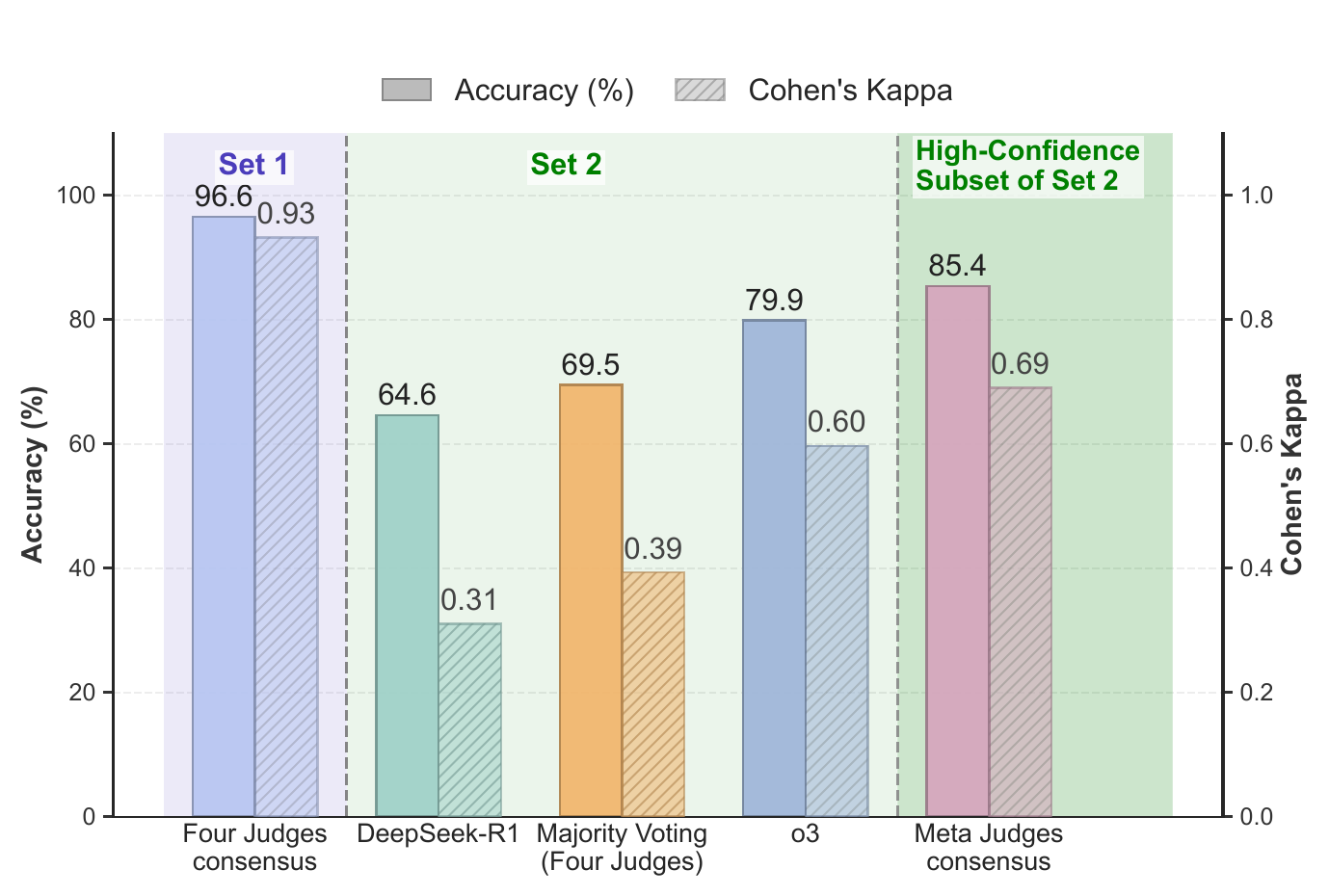}
     \caption{Preliminary experiments and results on human reference set.}
     \label{fig:raf_effect}
 \end{figure}

% ----------------

\subsubsection{Rubric Arbitration Framework}

We first evaluate a range of candidate judge models on reference set. Considering overall performance and practical trade-offs, we select four high-performance models spanning multiple model families as base judges. We then compare different labeling strategies built on these base judges. Different model performance and selection strategy are in Appendix~\ref{app:model_acc_on_human}.

As shown in Figure~\ref{fig:raf_effect}, when four base judges unanimously agree, their consensus achieves 96.6\% accuracy ($\kappa$=0.93). 
However, for disputed cases, majority voting yields only 69.5\% ($\kappa$=0.39), and even the best-performing single judge (o3) 
achieves just 79.9\% ($\kappa$=0.60).
Motivated by prior work showing that collaboration and meta-judging can improve evaluation reliability \cite{qian2025enhancing,wu2025meta}, we introduce a meta-review stage in which two meta-judges assess the rationales from multiple base judges and make their final judgment. We further enforce a consensus-based judgment rule to ensure high-quality labels. This strategy raises accuracy to 85.4\% ($\kappa$=0.69) on disputed cases.

Based on these findings, we propose the \textbf{Rubric Arbitration Framework (RAF)}, a three stage pipeline prioritizing label reliability over coverage—ambiguous instances are discarded rather than force-labeled. See Appendix \ref{app:case_study} for case study.

% -------

\paragraph{Coarse-grained Filtering}
Given the large number of rubrics, evaluating each one individually is costly. In this stage, four base judges evaluate the full rubric checklist for each 
instruction--response pair in a single pass. Rubrics with unanimous agreement are 
discarded; only disputed ones proceed to finer-grained stages. This procedure substantially alleviates the annotation burden in later stages

\begin{table}[t]
\centering
\small
\setlength{\tabcolsep}{4pt}
\begin{tabular}{llrrr}
\toprule
\textbf{Category} & \textbf{Benchmark} & \textbf{Easy} & \textbf{Hard} & \textbf{Total} \\
\midrule
\multirow{4}{*}{Constrained} 
  & InfoBench\_hard & 72 & 47 & 119 \\
  & ComplexBench    & 54 & 46 & 100 \\
  & CFBench         & 54 & 71 & 125 \\
  & AdvancedIF      & 188 & 152 & 340 \\
  & \textit{Subtotal} & 368 & 316 & 684 \\
\midrule
Compositional & ComplexBench & 130 & 110 & 240 \\
\midrule
\multirow{3}{*}{Multi-Turn} 
  & StructFlowBench & 186 & 95 & 281 \\
  & AdvancedIF      & 477 & 336 & 813 \\
  & \textit{Subtotal} & 663 & 431 & 1,094 \\
\midrule
\multirow{3}{*}{System} 
  & SysBench   & 160 & 217 & 377 \\
  & AdvancedIF & 713 & 378 & 1,091 \\
  & \textit{Subtotal} & 873 & 595 & 1,468 \\
\midrule
\multicolumn{2}{l}{\textbf{Total}} & \textbf{2,034} & \textbf{1,452} & \textbf{3,486} \\
\bottomrule
\end{tabular}
\caption{Statistics of \textsc{\textbf{RubricEval}} by instruction category 
and source benchmark.}
\label{tab:rubriceval_short_stats}
\end{table}

\begin{figure}[t]
    \centering
    \includegraphics[width=\linewidth,height=0.25\textheight,keepaspectratio]{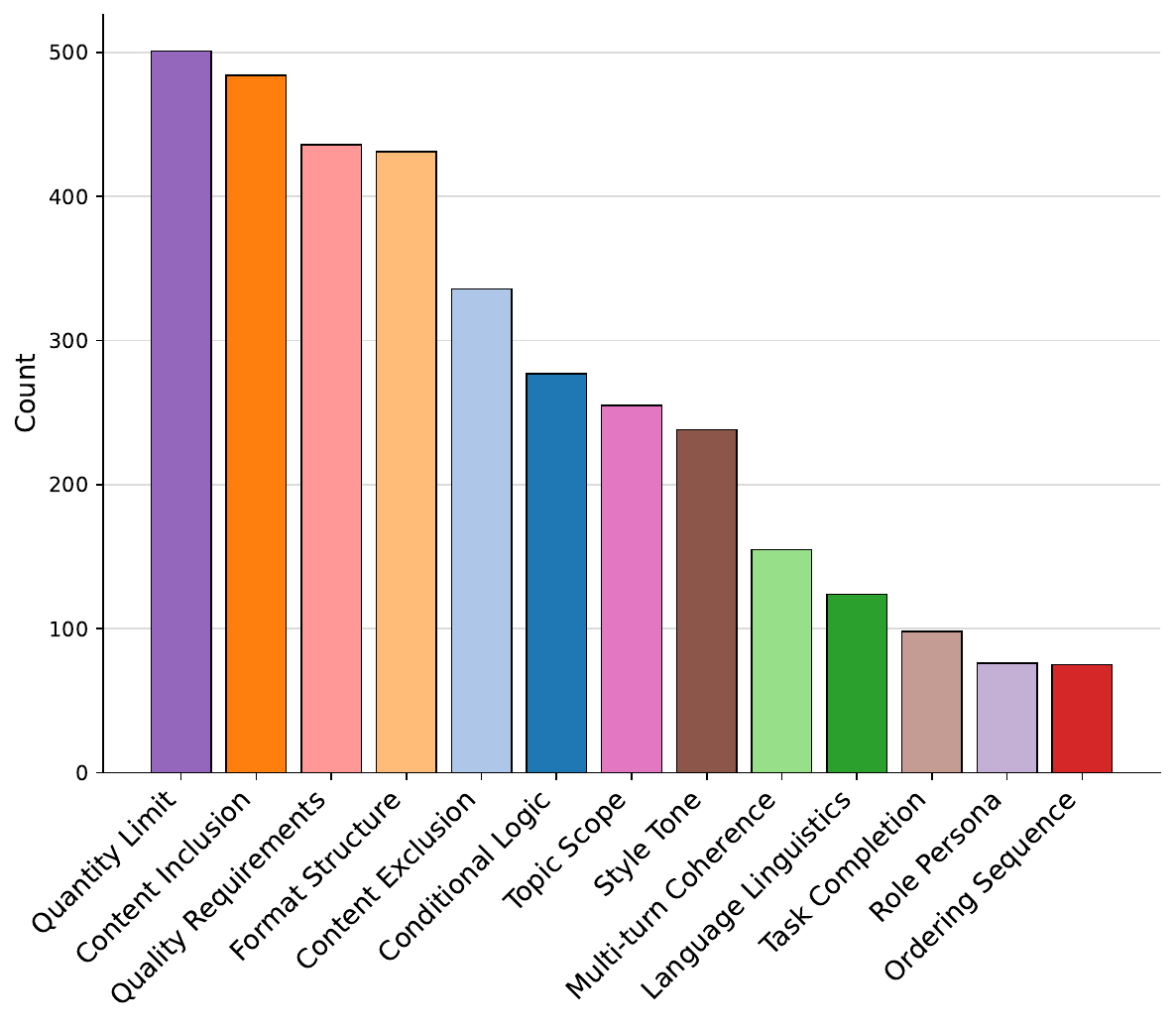}
    \caption{Distribution of instances across rubric categories in our taxonomy. }
    \label{fig:two_rubric_statis}
\end{figure}

\paragraph{Fine-grained Re-evaluation}
For disputed rubrics, we perform targeted rubric-level re-evaluation. Four base judge evaluates every disputed rubric in a single pass, providing both a judgment and supporting rationale.
Rubrics reaching unanimous agreement form \textsc{RubricEval-Easy}; others proceed to arbitration.
This stage forces judges focus evaluation on every rubric, reducing cross-rubric interference.

\paragraph{Meta-Judges Arbitration}
For persistently disputed rubrics, we invoke two meta-judges with strong reasoning capabilities.\footnote{We use OpenAI o3 and DeepSeek-R1 as meta-judges.} Two meta-judges arbitrate 
by reviewing the base judges' rationales and render independent judgments. 
When both agree, their consensus forms \textsc{RubricEval-Hard}; others are discarded.

% -----------
% todo：质量控制+对抛弃的rubric进行分析，说明并不是只抛弃了纯难的+Analysis of Discarded Rubrics，总结rubric经验
\subsection{Human Validation}

To verify the quality of our RAF-annotated labels, we conduct 
a human validation on a random sample of 160 rubric instances 
across subsets and instruction types. Two annotators independently 
label each instance, resolving disagreements through discussion.

Human-RAF agreement reaches 85.0\% accuracy with Cohen's $\kappa$ 
= \textbf{0.702}, indicating substantial agreement. This confirms that RAF reliably approximates 
human judgment and can serve as trustworthy ground truth for meta-evaluation.

\begin{table*}[t]
\centering
\small
\setlength{\tabcolsep}{3.5pt}
\resizebox{\textwidth}{!}{
\begin{tabular}{lcc|cc|cc|cc|cc}
\toprule
\multirow{2}{*}{\textbf{Model}} 
& \multicolumn{2}{c|}{\textbf{Constrained}} 
& \multicolumn{2}{c|}{\textbf{Compositional}} 
& \multicolumn{2}{c|}{\textbf{Multi-turn}} 
& \multicolumn{2}{c|}{\textbf{System}} 
& \multicolumn{2}{c}{\textbf{Overall}} \\
\cmidrule(lr){2-3} \cmidrule(lr){4-5} \cmidrule(lr){6-7} \cmidrule(lr){8-9} \cmidrule(lr){10-11}
 & \textbf{BAcc} & \textbf{mF1} 
 & \textbf{BAcc} & \textbf{mF1} 
 & \textbf{BAcc} & \textbf{mF1} 
 & \textbf{BAcc} & \textbf{mF1} 
 & \textbf{BAcc} & \textbf{mF1} \\
\midrule
% --- 下面内容保持原样 ---
\rowcolor{gray!15}
\multicolumn{11}{l}{\cellcolor{gray!15}\textit{\textsc{RubricEval-Easy}}} \\
\midrule
Llama-3.1-8B-Instruct & 65.71 & 61.18 & 57.52 & 48.21 & 65.80 & 58.25 & 66.12 & 64.59 & 63.79 & 58.06 \\ 
Llama-3.3-70B-Instruct & 80.48 & 81.83 & 81.65 & 83.42 & 82.82 & 83.83 & 86.83 & 87.24 & 82.94 & 84.08 \\ 
Qwen2.5-7B-Instruct & 63.70 & 63.05 & 69.85 & 62.46 & 60.06 & 55.04 & 67.11 & 65.89 & 65.18 & 61.61 \\ 
Qwen2.5-32B-Instruct & 75.78 & 76.26 & 75.64 & 70.01 & 75.72 & 76.96 & 79.38 & 79.17 & 76.63 & 75.60 \\ 
QwQ-32B & 85.26 & 85.67 & 79.17 & 73.80 & 77.00 & 77.12 & 80.29 & 80.38 & 80.43 & 79.24 \\ 
Qwen3-8B & 79.22 & 80.51 & 84.06 & 81.68 & 83.40 & 83.65 & 80.27 & 80.74 & 81.74 & 81.65 \\ 

\midrule
Qwen3-235B-A22B-2507 & \underline{87.24} & \textbf{88.56} & \textbf{93.98} & \textbf{92.44} & \underline{90.59} & \textbf{90.84} & 87.65 & 87.81 & \textbf{89.87} & \textbf{89.91} \\
gpt-oss-120b & \textbf{89.84} & \underline{87.24} & \underline{86.24} & 81.65 & \textbf{91.29} & \underline{89.65} & \textbf{90.83} & \textbf{90.02} & \underline{89.55} & \underline{87.14} \\
GPT-4o-2024-11-20 & 82.15 & 80.88 & 85.34 & 81.42 & 85.80 & 82.77 & 84.33 & 83.37 & 84.41 & 82.11 \\
o3-mini & 85.66 & 84.67 & 85.26 & \underline{84.63} & 88.63 & 89.40 & \underline{89.12} & \underline{89.15} & 87.17 & 86.96 \\

\midrule
\rowcolor{gray!15}
\multicolumn{11}{l}{\cellcolor{gray!15}\textit{\textsc{RubricEval-Hard}}} \\
\midrule
Qwen3-235B-A22B-2507 & 62.68 & 53.31 & 60.95 & 49.37 & 69.28 & 61.80 & 62.48 & 57.28 & 63.85 & 55.44 \\
gpt-oss-120b & 80.00 & 76.81 & 61.83 & 54.78 & 82.32 & 79.74 & 79.39 & 78.57 & 75.89 & 72.48 \\
GPT-4o-2024-11-20 & 48.51 & 41.04 & 67.99 & 53.27 & 57.85 & 56.20 & 49.54 & 48.21 & 55.97 & 49.68 \\
o3-mini & 71.88 & 64.78 & 69.97 & 55.86 & 76.63 & 70.03 & 62.48 & 57.28 & 70.24 & 61.99 \\

\midrule
GPT-4.1 & 64.91 & 55.67 & 52.92 & 44.60 & 74.39 & 71.24 & 60.09 & 57.31 & 63.08 & 57.21 \\
GPT-5.1 & 71.90 & 69.48 & 74.92 & 63.43 & 72.26 & 71.75 & 70.03 & 68.27 & 72.28 & 68.23 \\
o3    & \textbf{87.35} & \textbf{82.22} & 80.97 & 67.09 & \textbf{88.50} & \textbf{87.20} & \textbf{82.42} & \textbf{79.89} & \textbf{84.81} & \underline{79.10} \\
Claude-Sonnet-4.5 & 58.70 & 50.63 & 52.04 & 39.39 & 59.75 & 55.76 & 52.11 & 47.88 & 55.65 & 48.41 \\
Gemini-3-Flash & 77.91 & 75.30 & \textbf{84.43} & \underline{75.18} & 82.43 & 81.06 & 74.93 & 76.14 & 79.93 & 76.92 \\
Gemini-3-Pro & \underline{81.56} & \underline{78.08} & \underline{83.93} & \textbf{76.22} & \underline{87.77} & \underline{84.53} & \underline{78.89} & \underline{79.22} & \underline{83.04} & \textbf{79.51} \\
Deepseek-v3.2 & 53.48 & 50.08 & 59.85 & 52.28 & 64.41 & 60.90 & 58.95 & 55.67 & 59.17 & 54.73 \\
Deepseek-r1-0528 & 68.62 & 62.41 & 80.47 & 66.13 & 71.92 & 66.76 & 73.18 & 69.95 & 73.55 & 66.31 \\
\bottomrule
\end{tabular}
}
\caption{\label{tab:main_results}Main results on \textsc{RubricEval}. We report performance on the \textsc{Easy} (top) and \textsc{Hard} (bottom) splits of \textsc{RubricEval} across four instruction types and Overall. Each setting is evaluated with balanced accuracy (\textbf{BAcc}) and macro-F1 (\textbf{mF1}). \textbf{Bold} indicates the best score in each column within the same split, and \underline{underline} indicates the second-best.}
\end{table*}

% \begin{figure}[t]
%      \centering
%      \includegraphics[width=\linewidth]{pics/rubric_scatter_p30.pdf}
%      \caption{t-SNE visualization of rubric instances in the embedding space, colored by rubric category. Several categories form relatively compact clusters—e.g., [Multi-turn Coherence] (\textit{light green}), [Quantity Limit] (\textit{[purple]}), and [Format Structure] (\textit{[light orange]})—indicating consistent patterns within these rubric types.}
%      \label{fig:rubric_scatter_p30}
% \end{figure}
 
% \begin{figure}[t]
%      \centering
%      \includegraphics[width=\linewidth]{pics/rubric_count_bar.pdf}
%      \caption{Distribution of instances across rubric categories in our taxonomy, exhibiting a long-tail pattern reflecting a natural distribution of real-world tasks.}
%      \label{fig:rubric_count_bar}
% \end{figure}

\subsection{Dataset Statistics}
Table~\ref{tab:rubriceval_short_stats} summarizes the overall statistics of our constructed \textsc{RubricEval}. In total, the benchmark contains 1,989 instructions and 3,486 rubric-level instances, including 2,034 \textsc{Easy} and 1,452 \textsc{Hard} instances.  A detailed breakdown by source benchmark is provided in Appendix \ref{app:data_detailed_statis}.

To support fine-grained analysis of rubric-level judge performance, we construct a rubric taxonomy for \textsc{RubricEval}. This rubric taxonomy has \textbf{13} fine-grained categories, organized into 4 high-level dimensions: \textbf{Content}, \textbf{Form}, \textbf{Quality}, and \textbf{Style}. This taxonomy helps us view finer-grained rubric distributions. 

As shown in Figure \ref{fig:two_rubric_statis}, the distribution of instances exhibits a long-tail pattern reflecting a natural distribution of real-world tasks. 
And Appendix \ref{app:t-sne visual} show the t-SNE visualization of rubric instances.

The detailed category definitions and categorization procedure are provided in Appendix~\ref{app:rubric_taxonomy}. Appendix~\ref{app:rubric_statistics_pie} further summarizes the distribution by high-level dimensions (and their proportions).

% \subsection{Analysis of Discarded Rubrics}

\section{Experiments}
\subsection{Experimental Setup}

\paragraph{Metrics.}
Rubric-level judging is a binary classification task. We report \textbf{Balanced Accuracy} and \textbf{Macro F1} to account for class imbalance. 

\paragraph{Protocol.}
When conducting evaluation, we ask judges to first provide a rationale, then give the final judgment. We think this protocol better reflects the judge's true evaluation capability. We follow the original evaluation prompting guidelines provided by the corresponding source benchmarks. Prompt for evaluating \textit{Constrained} rubric is in Appendix \ref{app:prompts}.

\paragraph{Evaluated Models.}
We evaluate a diverse set of judge models covering both open-source and proprietary LLMs, spanning multiple model families and parameter scales. We report results on both the \textsc{Easy} and \textsc{Hard} splits, with an overlapping subset of models evaluated on both for direct comparison.

\subsection{Main Results}

% Section 5 Analysis
\newcommand{\cmark}{\ding{51}} % ✓
\newcommand{\xmark}{\ding{55}} % ✗

\begin{table*}[t]
    \centering
    \small
    \setlength{\tabcolsep}{4.5pt}
    % 定义绿色delta样式
    \definecolor{gaingreen}{RGB}{0, 150, 50}  % 自定义亮绿色
    \newcommand{\gain}[1]{\textcolor{gaingreen}{\scriptsize{(+#1)}}}
    \begin{tabular}{llcccccccccc}
    \toprule
    \multirow{2}{*}{\textbf{Granularity}} 
    & \multirow{2}{*}{\textbf{Reasoning}}
    & \multicolumn{2}{c}{\textbf{Constrained}} 
    & \multicolumn{2}{c}{\textbf{Compositional}} 
    & \multicolumn{2}{c}{\textbf{Multi-turn}} 
    & \multicolumn{2}{c}{\textbf{System}} 
    & \multicolumn{2}{c}{\textbf{Overall}} \\
    \cmidrule(lr){3-4} \cmidrule(lr){5-6} \cmidrule(lr){7-8} \cmidrule(lr){9-10} \cmidrule(lr){11-12}
    & & \textbf{Qwen} & \textbf{GPT}
    & \textbf{Qwen} & \textbf{GPT}
    & \textbf{Qwen} & \textbf{GPT}
    & \textbf{Qwen} & \textbf{GPT}
    & \textbf{Qwen} & \textbf{GPT}\\
    \midrule
    \multirow{2}{*}{Rubric-level}    
        & \xmark & 62.55 & 69.99 & 75.62 & 78.37 & 68.09 & 74.62 & 69.55 & 78.89 & 68.95 & 75.47 \\
        & \cmark & 74.19 & 82.90 & 83.17 & 80.83 & 76.80 & 83.48 & 75.34 & 81.45 & \textbf{77.38}\gain{8.4} & \textbf{82.17}\gain{6.7} \\
    \midrule
    \multirow{2}{*}{Checklist-level} 
        & \xmark & 54.06 & 60.80 & 72.38 & 68.21 & 53.39 & 56.11 & 63.98 & 68.48 & 60.95 & 63.40 \\
        & \cmark & 66.71 & 69.59 & 77.06 & 74.60 & 60.90 & 57.88 & 74.94 & 79.67 & 69.90\gain{9.0} & 70.44\gain{7.0} \\
    \bottomrule
    \end{tabular}
    \caption{Comparison of evaluation paradigms on a subset of \textsc{RubricEval}, sampled from both \textsc{Easy} and \textsc{Hard} subsets. We vary granularity and reasoning during evaluation, reporting Balanced Accuracy (\textbf{BAcc}) for Qwen (Qwen2.5-32B-Instruct) and GPT (GPT-4.1). \textcolor{gaingreen}{Green values} show improvement from reasoning. \textbf{Bold} indicates the best score in Overall column. Results on Easy and Hard subsets are in Appendix~\ref{app:paradigms_ez_hd}.}
    \label{tab:paradigm_mixed}
\end{table*}

Table~\ref{tab:main_results} reports the main results on \textsc{RubricEval}.

\paragraph{Overall Performance.}
The results reveal a wide performance spectrum across evaluated models. On the \textsc{Easy} subset, small open-source models such as Qwen2.5-7B-Instruct achieves only around 65\% balanced accuracy, while stronger models like Qwen3-235B and gpt-oss-120b reach around 90\%. On the \textsc{Hard} split, even commercial models struggle considerably. For instance, GPT-4o achieves merely 55.97\% balanced accuracy, and Claude-Sonnet-4.5 reaches 55.65\%, indicating that hard rubric cases remain challenging even for strong LLMs. These findings underscore the necessity of rubric-level meta-evaluation: \textit{rubric-level judging remains far from solved}. 
Practically, deploying small open-source models as judges\cite{qin2025incentivizing} may produce noisy or misleading signals in applications like rubric-based RL. Meanwhile, GPT-4o, the widely-adopted evaluator in instruction-following benchmarks\cite{wen2024benchmarking,li2025structflowbench}, may introduce systematic biases, potentially affecting the reliability of the reported scores.

\paragraph{From \textsc{Easy} to \textsc{Hard}: A Significant Performance Gap.}
We evaluate the same four models on both \textsc{Easy} and \textsc{Hard} subsets\footnote{We evaluate these four models: Qwen3-235B-A22B-2507, gpt-oss-120b, GPT-4o, and o3-mini}, enabling a direct comparison of subset difficulty. Consistently, all four exhibit substantial performance drop from \textsc{Easy} to \textsc{Hard}: GPT-4o declines by 28.4 BAcc (84.41\% $\rightarrow$ 55.97\%), Qwen3-235B by 26.0 points (89.87\% $\rightarrow$ 63.85\%), and even the relatively robust gpt-oss-120b drops by 13.7 points. 
This performance degradation confirms that our data construction pipeline produces two practical subsets that vary in difficulty. \textsc{Hard} subset genuinely captures challenging cases. The two-tier design also facilitates more fine-grained evaluation across a broader set of judges.

\paragraph{Performance Varies Across Instruction Types.}
Judge performance also varies across instruction categories. This indicates that rubric verification difficulty depends strongly on the type of the underlying instruction. \textit{Compositional} instructions prove most challenging, with most models showing their lowest mF1 on this type. This is likely because judges are required to accurately parse the underlying structure and ground each rubric to specific parts of the response, which is more error-prone than checking surface-level constraints.
Conversely, \textit{Multi-turn} instructions tend to be easier, possibly because conversational history provides additional cues for rubric verification. \textit{Constrained} and \textit{System} instructions show moderate difficulty, though some models underperform notably.

\section{Analysis}
In this section, we study different evaluation paradigms that vary in granularity and reasoning, as well as common error patterns.

\subsection{Does Evaluation Paradigm Matter?}
Table~\ref{tab:paradigm_mixed} compares four evaluation paradigms along two dimensions: \textbf{granularity} and \textbf{reasoning}. For granularity, checklist-level evaluates all rubrics in a single pass, 
while rubric-level verifies each rubric independently with a separate call. 
For reasoning, we compare direct judgment versus generating 
a rationale before the final verdict.

\paragraph{Rubric-Level Evaluation is More Accurate.}
As shown in Table~\ref{tab:paradigm_mixed}, rubric-level evaluation consistently outperforms checklist-level evaluation across both models and all instruction types. 
With reasoning enabled, rubric-level achieves 77.38\% (Qwen) and 82.17\% (GPT) BAcc, while checklist-level achieves only 69.90\% and 70.44\%—a gap of 7–12 points. This pattern holds across all instruction types and on both \textsc{Easy} and \textsc{Hard} 
subsets (see Appendix~\ref{app:paradigms_ez_hd}).

\paragraph{Reasoning Consistently Helps.}
Explicit reasoning also significantly enhances judging accuracy. Across both granularity settings and all model types, enabling reasoning consistently leads to performance gains. Specifically, in the rubric-level setting, Qwen and GPT achieve absolute improvements of 8.4\% and 6.7\% in BAcc, respectively. Similar trends are observed in the checklist-level setting, with gains of 9.0\% and 7.0\%. 

 \begin{figure}[t]
     \centering
     \includegraphics[width=\linewidth]{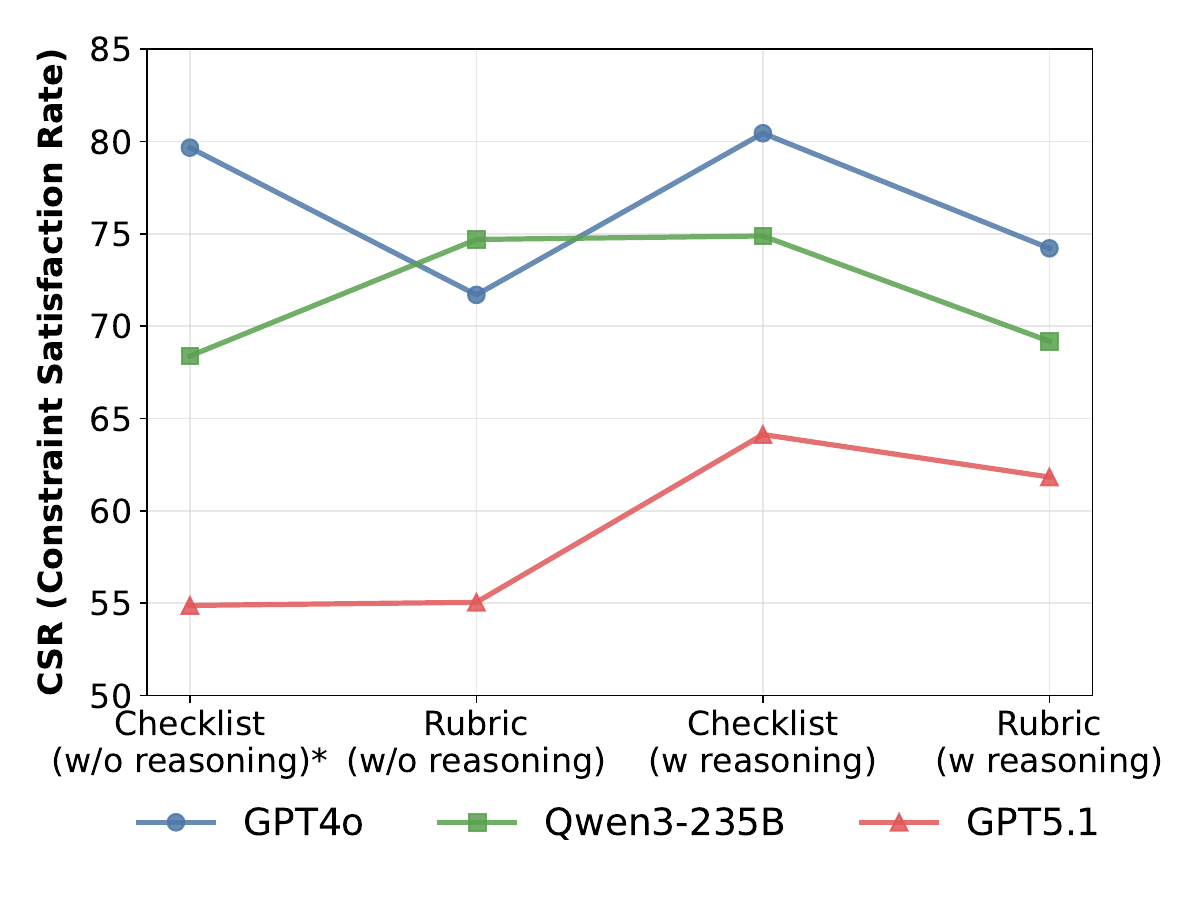}
     \caption{Inter-judge analysis on CFBench. Judge variance decreases from vanilla (*) to rubric-level with reasoning.}
     \label{fig:cfbench_lines}
 \end{figure}
 
\subsection{Trade-offs}
For the first finding, a likely explanation is that checklist-level evaluation forces judges to verify multiple rubrics in a single pass, increasing cognitive load and the risk of missing individual rubrics. 
Rubric-level evaluation isolates each decision, reducing interference and improving precision.

For the second finding, reasoning likely helps by forcing judges to ground their decisions in evidence rather than relying on intuition, thereby reducing unreliable judgments.

However, both rubric-level evaluation and reasoning come at a cost. 
Rubric-level evaluation requires a separate API call for each rubric, significantly increasing latency and expense. Reasoning further adds to output token costs. This creates a \textbf{reliability--efficiency trade-off}: checklist-level without reasoning is fast and cheap but less accurate, while rubric-level with reasoning is more reliable but costlier. Some existing benchmarks and reward methods adopt 
the former for efficiency—our results suggest this may compromise evaluation reliability.

\subsection{Inter-Judge Analysis}
We further investigate whether evaluation paradigm affects 
inter-judge consistency. Using CFBench\cite{zhang2025cfbench} as a testbed, we evaluate responses from Qwen2.5-7B-Instruct with three judges of varying performance levels on our benchmark (GPT-4o, Qwen3-235B, and GPT-5.1) under four evaluation paradigms.

As shown in Figure~\ref{fig:cfbench_lines}, the vanilla evaluation paradigm (checklist-level without reasoning) exhibits substantial inter-judge variance: CSR scores range from 55\% (GPT-5.1) to 80\% (GPT-4o)—a gap of \textbf{25} points for the same model responses. This suggests that judge selection alone can dramatically affect benchmark scores and potentially lead to conflicting conclusions about model performance.

As we move toward more fine-grained and reasoning-augmented paradigms, inter-judge variance decreases. With rubric-level evaluation and reasoning, the three judges converge noticeably: scores range from 62\% to 74\%, reducing the gap to \textbf{12} points. However, non-trivial differences still remain, reflecting inherent capability gaps among judges. This suggests that rubric-level evaluation with reasoning improves both accuracy and inter-judge consistency, but cannot fully eliminate variance introduced by judge capability differences, 
highlighting the importance of our rubric-level meta-evaluation efforts.

%\subsection{Intra-Model}

% Rubric Type Performance Table for ACL Paper (single column)
% 4 models across 4 instruction types (merged)
% Underlined values are below the dimension average for that model

\begin{table}[t]
    \centering
    \small
    \setlength{\tabcolsep}{2pt}
    \begin{tabular}{@{}l cccc@{}}
    \toprule
    \textbf{Rubric Type} & \textbf{Qwen3} & \textbf{gpt-oss} & \textbf{GPT-4o} & \textbf{o3-mini} \\
    \midrule
    \textbf{Content} & 76.9 & 86.1 & 72.2 & 81.0 \\
    \quad Content Inclusion & 77.1 & 86.8 & 72.3 & \underline{80.6} \\
    \quad Content Exclusion & 80.4 & 87.8 & \underline{71.4} & 83.9 \\
    \quad Topic Scope & \underline{72.9} & \underline{83.5} & 72.9 & \underline{78.4} \\
    \midrule
    \textbf{Form} & 77.6 & 87.5 & 67.0 & 87.3 \\
    \quad Quantity Limit & 78.6 & \underline{86.6} & 68.7 & 88.6 \\
    \quad Format Structure & \underline{75.9} & \underline{86.1} & \underline{66.8} & \underline{84.9} \\
    \quad Ordering Sequence & 80.0 & 94.7 & \underline{61.3} & 89.3 \\
    \midrule
    \textbf{Quality} & 75.6 & 84.5 & 70.8 & 76.6 \\
    \quad Quality Requirements & \underline{74.8} & \underline{83.2} & 72.5 & \underline{75.2} \\
    \quad Conditional Logic & 78.0 & 84.8 & 71.8 & 80.9 \\
    \quad Task Completion & \underline{73.5} & 86.6 & \underline{66.3} & \underline{72.4} \\
    \midrule
    \textbf{Style} & 79.2 & 86.6 & 74.7 & 78.5 \\
    \quad Style Tone & 79.4 & 87.0 & 79.0 & 79.0 \\
    \quad Language Linguistics & \underline{75.8} & 89.4 & \underline{70.2} & 83.9 \\
    \quad Multi-turn Coherence & 91.0 & 89.0 & 75.6 & \underline{74.2} \\
    \quad Role Persona & \underline{71.1} & \underline{81.6} & \underline{72.0} & \underline{77.6} \\
    \midrule
    \textbf{Overall} & 77.5 & 86.2 & 71.2 & 81.4 \\
    \bottomrule
    \end{tabular}
    \caption{Judges accuracy (\%) by rubric type. \underline{Underlined} values are below the dimension average for that model.}
    \label{tab:error_analysis}
\end{table}

\subsection{Error Analysis}

Table~\ref{tab:error_analysis} presents the accuracy of four judges\footnotetext{Qwen3 refers to Qwen3-235B-A22B-Instruct-2507, gpt-oss refers to gpt-oss-120b, and GPT-4o refers to GPT-4o-2024-11-20.} across fine-grained rubric types in \textsc{RubricEval}, grouped into four high-level dimensions.

\paragraph{Common Failure Modes.}
We identify five rubric types where most judges underperform: 
\textit{\textbf{Topic Scope}}, \textit{\textbf{Format Structure}}, \textit{\textbf{Quality Requirements}}, 
\textit{\textbf{Task Completion}}, and \textit{\textbf{Role Persona}}. These rubrics 
typically require strict evidence checking or involve subjective interpretation.

\textit{Format Structure} and \textit{Role Persona} are consistently 
difficult across all judges—the former reveals that format structure verification is relatively hard for llm judges and may benefit from rule-based verification methods. While the latter indicates that persona maintenance remains ambiguous for judges to assess, where correctness is not always clearly defined and borderline cases may exist.
The other three types (\textit{Topic Scope}, \textit{Quality Requirements}, \textit{Task Completion}) all lack clear-cut 
criteria, making consistent judgment difficult.

\paragraph{Model-Specific Observations.}
We also observe clear model-specific strengths and weaknesses. GPT-4o performs poorly on \textit{Form} rubrics (67.0\%), especially on \textit{Ordering/Sequence} (61.3\%), suggesting difficulty in verifying strict ordering requirements. In contrast, Qwen3 performs strongly on \textit{Multi-turn Coherence} (91.0\%), indicating better handling of dialogue consistency across turns. The gpt-oss judge shows the most balanced performance across dimensions overall, although it still underperforms on \textit{Role Persona}, which remains challenging across models.

\section{Conclusion}

We present \textsc{\textbf{RubricEval}}, the first rubric-level meta-evaluation benchmark for instruction following, covering four instruction categories with \textsc{Easy} and \textsc{Hard} splits. We design and use the Rubric Arbitration Framework (RAF) to produce high-confidence labels at scale.
Our experiments reveal that rubric-level judging remains challenging. Even widely adopted judges like GPT-4o and Claude-4.5 struggle on hard instances, raising concerns about current rubric-based evaluation practices. We also find that rubric-level evaluation outperforms checklist-level evaluation, explicit reasoning improves judging accuracy, and both together enhance inter-judge consistency. Through error analysis with our rubric taxonomy, we identify common failure modes, providing guidance for future judge development and benchmark design.
We hope \textsc{RubricEval} serves as a foundation for developing more reliable LLM judges for instruction following, ultimately advancing trustworthy evaluation in both research and practice.

\section*{Limitations}

Our work has several limitations: (1) \textsc{RubricEval} focuses on four main instruction categories, which may not fully cover all instruction-following scenarios in practice. Other instruction types, such as agent-related or domain-specific instructions, are not included.
(2) The Rubric Arbitration Framework (RAF) relies on LLM judges and reasoning models to produce high-confidence reference labels. Although human validation shows high agreement with RAF labels, the remaining cases may still contain annotation noise. Additionally, rubrics that fail to reach consensus between meta-judges are discarded to prioritize label quality. While the resulting benchmark remains sufficiently large and discriminative, some genuinely hard cases may still be excluded.
(3) We focus on rubric-level binary judgments, which is the most common setting in current benchmarks. Other evaluation formats, such as Likert-scale ratings or comparative judgments, are beyond the scope of our work.

\bibliography{custom}

@article{zhou2023instruction,
  title={Instruction-following evaluation for large language models},
  author={Zhou, Jeffrey and Lu, Tianjian and Mishra, Swaroop and Brahma, Siddhartha and Basu, Sujoy and Luan, Yi and Zhou, Denny and Hou, Le},
  journal={arXiv preprint arXiv:2311.07911},
  year={2023}
}

@article{pyatkin2025generalizing,
  title={Generalizing Verifiable Instruction Following},
  author={Pyatkin, Valentina and Malik, Saumya and Graf, Victoria and Ivison, Hamish and Huang, Shengyi and Dasigi, Pradeep and Lambert, Nathan and Hajishirzi, Hannaneh},
  journal={arXiv preprint arXiv:2507.02833},
  year={2025}
}

@article{he2024multi,
  title={Multi-if: Benchmarking llms on multi-turn and multilingual instructions following},
  author={He, Yun and Jin, Di and Wang, Chaoqi and Bi, Chloe and Mandyam, Karishma and Zhang, Hejia and Zhu, Chen and Li, Ning and Xu, Tengyu and Lv, Hongjiang and others},
  journal={arXiv preprint arXiv:2410.15553},
  year={2024}
}

@inproceedings{he2024can,
  title={Can large language models understand real-world complex instructions?},
  author={He, Qianyu and Zeng, Jie and Huang, Wenhao and Chen, Lina and Xiao, Jin and He, Qianxi and Zhou, Xunzhe and Liang, Jiaqing and Xiao, Yanghua},
  booktitle={Proceedings of the AAAI Conference on Artificial Intelligence},
  volume={38},
  number={16},
  pages={18188--18196},
  year={2024}
}

@article{qin2024infobench,
  title={Infobench: Evaluating instruction following ability in large language models},
  author={Qin, Yiwei and Song, Kaiqiang and Hu, Yebowen and Yao, Wenlin and Cho, Sangwoo and Wang, Xiaoyang and Wu, Xuansheng and Liu, Fei and Liu, Pengfei and Yu, Dong},
  journal={arXiv preprint arXiv:2401.03601},
  year={2024}
}

@article{wen2024benchmarking,
  title={Benchmarking complex instruction-following with multiple constraints composition},
  author={Wen, Bosi and Ke, Pei and Gu, Xiaotao and Wu, Lindong and Huang, Hao and Zhou, Jinfeng and Li, Wenchuang and Hu, Binxin and Gao, Wendy and Xu, Jiaxing and others},
  journal={Advances in Neural Information Processing Systems},
  volume={37},
  pages={137610--137645},
  year={2024}
}

@inproceedings{zhang2025cfbench,
  title={Cfbench: A comprehensive constraints-following benchmark for llms},
  author={Zhang, Tao and Zhu, Chenglin and Shen, Yanjun and Luo, Wenjing and Zhang, Yan and Liang, Hao and Yang, Fan and Lin, Mingan and Qiao, Yujing and Chen, Weipeng and others},
  booktitle={Proceedings of the 63rd Annual Meeting of the Association for Computational Linguistics (Volume 1: Long Papers)},
  pages={32926--32944},
  year={2025}
}

@article{malik2025rewardbench,
  title={RewardBench 2: Advancing Reward Model Evaluation},
  author={Malik, Saumya and Pyatkin, Valentina and Land, Sander and Morrison, Jacob and Smith, Noah A and Hajishirzi, Hannaneh and Lambert, Nathan},
  journal={arXiv preprint arXiv:2506.01937},
  year={2025}
}

@article{tan2024judgebench,
  title={Judgebench: A benchmark for evaluating llm-based judges},
  author={Tan, Sijun and Zhuang, Siyuan and Montgomery, Kyle and Tang, William Y and Cuadron, Alejandro and Wang, Chenguang and Popa, Raluca Ada and Stoica, Ion},
  journal={arXiv preprint arXiv:2410.12784},
  year={2024}
}

@article{zeng2023evaluating,
  title={Evaluating large language models at evaluating instruction following},
  author={Zeng, Zhiyuan and Yu, Jiatong and Gao, Tianyu and Meng, Yu and Goyal, Tanya and Chen, Danqi},
  journal={arXiv preprint arXiv:2310.07641},
  year={2023}
}

@inproceedings{liu2025reife,
  title={ReIFE: Re-evaluating instruction-following evaluation},
  author={Liu, Yixin and Shi, Kejian and Fabbri, Alexander Richard and Zhao, Yilun and Wang, Peifeng and Wu, Chien-Sheng and Joty, Shafiq and Cohan, Arman},
  booktitle={Proceedings of the 2025 Conference of the Nations of the Americas Chapter of the Association for Computational Linguistics: Human Language Technologies (Volume 1: Long Papers)},
  pages={12247--12287},
  year={2025}
}

@article{qin2025incentivizing,
  title={Incentivizing Reasoning for Advanced Instruction-Following of Large Language Models},
  author={Qin, Yulei and Li, Gang and Li, Zongyi and Xu, Zihan and Shi, Yuchen and Lin, Zhekai and Cui, Xiao and Li, Ke and Sun, Xing},
  journal={arXiv preprint arXiv:2506.01413},
  year={2025}
}

@article{peng2025verif,
  title={VerIF: Verification Engineering for Reinforcement Learning in Instruction Following},
  author={Peng, Hao and Qi, Yunjia and Wang, Xiaozhi and Xu, Bin and Hou, Lei and Li, Juanzi},
  journal={arXiv preprint arXiv:2506.09942},
  year={2025}
}

@article{ferraz2024llm,
  title={Llm self-correction with decrim: Decompose, critique, and refine for enhanced following of instructions with multiple constraints},
  author={Ferraz, Thomas Palmeira and Mehta, Kartik and Lin, Yu-Hsiang and Chang, Haw-Shiuan and Oraby, Shereen and Liu, Sijia and Subramanian, Vivek and Chung, Tagyoung and Bansal, Mohit and Peng, Nanyun},
  journal={arXiv preprint arXiv:2410.06458},
  year={2024}
}

@article{li2025verifybench,
  title={Verifybench: A systematic benchmark for evaluating reasoning verifiers across domains},
  author={Li, Xuzhao and Li, Xuchen and Hu, Shiyu and Guo, Yongzhen and Zhang, Wentao},
  journal={arXiv preprint arXiv:2507.09884},
  year={2025}
}

@article{ren2025step,
  title={Step-by-step mastery: Enhancing soft constraint following ability of large language models},
  author={Ren, Qingyu and Zeng, Jie and He, Qianyu and Liang, Jiaqing and Xiao, Yanghua and Zhou, Weikang and Sun, Zeye and Yu, Fei},
  journal={arXiv preprint arXiv:2501.04945},
  year={2025}
}

@article{ouyang2022training,
  title={Training language models to follow instructions with human feedback},
  author={Ouyang, Long and Wu, Jeffrey and Jiang, Xu and Almeida, Diogo and Wainwright, Carroll and Mishkin, Pamela and Zhang, Chong and Agarwal, Sandhini and Slama, Katarina and Ray, Alex and others},
  journal={Advances in neural information processing systems},
  volume={35},
  pages={27730--27744},
  year={2022}
}

@article{achiam2023gpt,
  title={Gpt-4 technical report},
  author={Achiam, Josh and Adler, Steven and Agarwal, Sandhini and Ahmad, Lama and Akkaya, Ilge and Aleman, Florencia Leoni and Almeida, Diogo and Altenschmidt, Janko and Altman, Sam and Anadkat, Shyamal and others},
  journal={arXiv preprint arXiv:2303.08774},
  year={2023}
}

@article{arora2025healthbench,
  title={Healthbench: Evaluating large language models towards improved human health},
  author={Arora, Rahul K and Wei, Jason and Hicks, Rebecca Soskin and Bowman, Preston and Qui{\~n}onero-Candela, Joaquin and Tsimpourlas, Foivos and Sharman, Michael and Shah, Meghan and Vallone, Andrea and Beutel, Alex and others},
  journal={arXiv preprint arXiv:2505.08775},
  year={2025}
}

@article{gunjal2025rubrics,
  title={Rubrics as rewards: Reinforcement learning beyond verifiable domains},
  author={Gunjal, Anisha and Wang, Anthony and Lau, Elaine and Nath, Vaskar and He, Yunzhong and Liu, Bing and Hendryx, Sean},
  journal={arXiv preprint arXiv:2507.17746},
  year={2025}
}

@article{huang2025reinforcement,
  title={Reinforcement learning with rubric anchors},
  author={Huang, Zenan and Zhuang, Yihong and Lu, Guoshan and Qin, Zeyu and Xu, Haokai and Zhao, Tianyu and Peng, Ru and Hu, Jiaqi and Shen, Zhanming and Hu, Xiaomeng and others},
  journal={arXiv preprint arXiv:2508.12790},
  year={2025}
}

@article{he2025advancedif,
  title={AdvancedIF: Rubric-Based Benchmarking and Reinforcement Learning for Advancing LLM Instruction Following},
  author={He, Yun and Li, Wenzhe and Zhang, Hejia and Li, Songlin and Mandyam, Karishma and Khosla, Sopan and Xiong, Yuanhao and Wang, Nanshu and Peng, Xiaoliang and Li, Beibin and others},
  journal={arXiv preprint arXiv:2511.10507},
  year={2025}
}

@article{qin2024sysbench,
  title={SysBench: Can Large Language Models Follow System Messages?},
  author={Qin, Yanzhao and Zhang, Tao and Shen, Yanjun and Luo, Wenjing and Sun, Haoze and Zhang, Yan and Qiao, Yujing and Chen, Weipeng and Zhou, Zenan and Zhang, Wentao and others},
  journal={arXiv preprint arXiv:2408.10943},
  year={2024}
}

@article{li2025structflowbench,
  title={Structflowbench: A structured flow benchmark for multi-turn instruction following},
  author={Li, Jinnan and Li, Jinzhe and Wang, Yue and Chang, Yi and Wu, Yuan},
  journal={arXiv preprint arXiv:2502.14494},
  year={2025}
}

@article{an2025ultraif,
  title={UltraIF: Advancing Instruction Following from the Wild},
  author={An, Kaikai and Sheng, Li and Cui, Ganqu and Si, Shuzheng and Ding, Ning and Cheng, Yu and Chang, Baobao},
  journal={arXiv preprint arXiv:2502.04153},
  year={2025}
}

@article{viswanathan2025checklists,
  title={Checklists are better than reward models for aligning language models},
  author={Viswanathan, Vijay and Sun, Yanchao and Ma, Shuang and Kong, Xiang and Cao, Meng and Neubig, Graham and Wu, Tongshuang},
  journal={arXiv preprint arXiv:2507.18624},
  year={2025}
}

@article{liu2025recast,
  title={RECAST: Strengthening LLMs' Complex Instruction Following with Constraint-Verifiable Data},
  author={Liu, Wenhao and Guo, Zhengkang and Xie, Mingchen and Xu, Jingwen and Huang, Zisu and Tian, Muzhao and Xu, Jianhan and Wu, Muling and Wang, Xiaohua and Lv, Changze and others},
  journal={arXiv preprint arXiv:2505.19030},
  year={2025}
}

@inproceedings{zhang2025iopo,
  title={Iopo: Empowering llms with complex instruction following via input-output preference optimization},
  author={Zhang, Xinghua and Yu, Haiyang and Fu, Cheng and Huang, Fei and Li, Yongbin},
  booktitle={Proceedings of the 63rd Annual Meeting of the Association for Computational Linguistics (Volume 1: Long Papers)},
  pages={22185--22200},
  year={2025}
}

@article{zhou2025evaluating,
  title={Evaluating judges as evaluators: The jetts benchmark of llm-as-judges as test-time scaling evaluators},
  author={Zhou, Yilun and Xu, Austin and Wang, Peifeng and Xiong, Caiming and Joty, Shafiq},
  journal={arXiv preprint arXiv:2504.15253},
  year={2025}
}

@article{wang2025light,
  title={Light-IF: Endowing LLMs with Generalizable Reasoning via Preview and Self-Checking for Complex Instruction Following},
  author={Wang, Chenyang and Wen, Liang and Jia, Shousheng and Zhang, Xiangzheng and Xu, Liang},
  journal={arXiv preprint arXiv:2508.03178},
  year={2025}
}

@article{lior2025wildifeval,
  title={Wildifeval: Instruction following in the wild},
  author={Lior, Gili and Yehudai, Asaf and Gera, Ariel and Ein-Dor, Liat},
  journal={arXiv preprint arXiv:2503.06573},
  year={2025}
}

@article{bai2024mt,
  title={Mt-bench-101: A fine-grained benchmark for evaluating large language models in multi-turn dialogues},
  author={Bai, Ge and Liu, Jie and Bu, Xingyuan and He, Yancheng and Liu, Jiaheng and Zhou, Zhanhui and Lin, Zhuoran and Su, Wenbo and Ge, Tiezheng and Zheng, Bo and others},
  journal={arXiv preprint arXiv:2402.14762},
  year={2024}
}

@article{zhang2025iheval,
  title={IHEval: Evaluating language models on following the instruction hierarchy},
  author={Zhang, Zhihan and Li, Shiyang and Zhang, Zixuan and Liu, Xin and Jiang, Haoming and Tang, Xianfeng and Gao, Yifan and Li, Zheng and Wang, Haodong and Tan, Zhaoxuan and others},
  journal={arXiv preprint arXiv:2502.08745},
  year={2025}
}

@article{qian2025enhancing,
  title={Enhancing LLM-as-a-judge via multi-agent collaboration},
  author={Qian, Yiyue and Zhang, Shinan and Zhou, Yun and Ding, Haibo and Socolinsky, Diego and Zhang, Yi},
  year={2025}
}

@inproceedings{wu2025meta,
  title={Meta-rewarding language models: Self-improving alignment with llm-as-a-meta-judge},
  author={Wu, Tianhao and Yuan, Weizhe and Golovneva, Olga and Xu, Jing and Tian, Yuandong and Jiao, Jiantao and Weston, Jason E and Sukhbaatar, Sainbayar},
  booktitle={Proceedings of the 2025 Conference on Empirical Methods in Natural Language Processing},
  pages={11548--11565},
  year={2025}
}

\appendix

\newpage
\section{Instruction Category Definitions}
\label{app:instruction_categories}

We collect instructions in \textsc{RubricEval} from four widely used categories. Below we provide detailed definitions for each category.

\paragraph{Constrained Instructions.}
Constrained instructions are single-turn instructions that contain 
multiple constraints that model must satisfy simultaneously during generation. 
For example, an instruction may require the response to simultaneously include specific content, follow a specified format, and output in a style.

This type of instruction is widely used in instruction-following evaluation, 
as it directly tests a model's ability to handle multiple requirements in parallel. 
Representative benchmarks include InfoBench~\cite{qin2024infobench}, CFBench~\cite{zhang2025cfbench}, TRACE~\cite{zhang2025iopo} and Wildifeval\cite{lior2025wildifeval}.

Evaluation difficulty for constrained instructions is moderate to high, as judges must verify each constraint independently 
while ensuring no constraint is overlooked.

\paragraph{Compositional Instructions.}
Compositional instructions contain complex topological structures 
with logical dependencies among constraints, such as conditional branches(Selection), sequential chains(Chain), and conjunctive relations(And). 

This type of instruction tests a model's ability to parse and execute logically structured requirements, which is essential for complex real-world tasks. ComplexBench~\cite{wen2024benchmarking} is the primary benchmark focusing on this instruction type.

Evaluation difficulty is high, as judges must correctly parse 
the underlying logical structure and ground each rubric 
to the corresponding part of the response.

\paragraph{Multi-turn Instructions.}
Multi-turn instructions involve conversational interactions 
spanning multiple dialogue turns. The model must maintain consistency, track context, and follow constraints that may evolve or accumulate across turns.

This type of instruction reflects realistic conversational AI scenarios, where users interact with models through extended dialogues. 
Related benchmarks include MT-Bench-101~\cite{bai2024mt} 
and StructFlowBench\cite{li2025structflowbench}.

Evaluation difficulty is moderate, as conversational history 
provides additional context for rubric verification. 
However, judges must correctly handle cross-turn references 
and ensure coherence throughout the conversation.

\paragraph{System Instructions.}
System instructions include a system prompt that defines 
the model's behavior, role, or constraints at the conversation level. 
The model is expected to strictly adhere to the system prompt 
throughout its responses.

This type of instruction is prevalent in deployed AI systems, 
where system prompts are used to customize model behavior 
for specific applications. Benchmarks such as 
SysBench~\cite{qin2024sysbench} and IHEval\cite{zhang2025iheval} focus on system-prompt following evaluation.

Evaluation difficulty varies depending on the specificity 
of the system prompt. Verifying adherence to abstract role 
definitions (e.g., ``act as a helpful assistant'') is harder 
than checking concrete constraints (e.g., ``always respond in JSON'').

\section{Statistics of the original instructions and rubrics}
\label{app:source_ins_rubs}
\begin{table}[H]
\centering
\small
\setlength{\tabcolsep}{2.5pt}
\begin{tabular}{lccccc}
\toprule
\textbf{Type} & \textbf{Benchmark} & \textbf{\#Inst.} & \textbf{\#Rub.} & \textbf{R/I} & \textbf{H.} \\
\midrule
\multirow{4}{*}{Constrained} & InfoBench\_hard & 228 & 1,453 & 6.37 & \checkmark \\
 & ComplexBench & 238 & 1,027 & 4.32 & \checkmark \\
 & CFBench & 243 & 1,035 & 4.26 & \checkmark \\
 & AdvancedIF & 243 & 1,816 & 7.47 & \checkmark \\
\midrule
Compositional & ComplexBench & 435 & 1,651 & 4.49 & \checkmark \\
\midrule
\multirow{2}{*}{Multi-Turn} & StructFlowBench & 643 & 1,775 & 2.76 & \checkmark \\
 & AdvancedIF & 736 & 4,478 & 6.08 & \checkmark \\
\midrule
\multirow{2}{*}{System} & SysBench & 1,000 & 2,478 & 2.48 & \checkmark \\
 & AdvancedIF & 507 & 4,972 & 9.81 & \checkmark \\
\midrule
\multicolumn{2}{l}{\textbf{Total}} & \textbf{4,273} & \textbf{20,685} & \textbf{4.84} & \checkmark \\
\bottomrule
\end{tabular}
\caption{Instruction sources and rubric statistics in \textsc{RubricEval}. Statistics are computed over the benchmark subsets used in our experiments. \#Inst.: instructions; \#Rub.: rubrics; R/I: rubrics per instruction; H.: human-crafted/verified.}

\label{tab:ins_rub_collection}
\end{table}

Table~\ref{tab:ins_rub_collection} summarizes the instruction and rubric sources used in \textsc{RubricEval}. We collect from multiple benchmarks across four instruction categories, totaling 4,273 instructions and 20,685 rubrics. All rubrics are human-crafted or human-verified.

\section{Rubric-based Evaluation}
\label{app:rubric-based_eval}
Rubric-based evaluation has been widely adopted across various 
domains beyond instruction following. For example, 
HealthBench~\cite{arora2025healthbench} employs rubric-level 
verification to evaluate medical question answering, and similar 
approaches have been applied to code generation, summarization, 
and other complex tasks. In this paradigm, complex evaluation 
criteria are decomposed into a set of fine-grained rubrics, each 
specifying a particular requirement. An LLM judge then verifies 
whether the response satisfies each rubric independently, and 
the results are aggregated into an overall score.

Beyond benchmarking, rubric-level judgments are increasingly 
used as supervision or reward signals in model 
training~\cite{gunjal2025rubrics,huang2025reinforcement,
peng2025verif,an2025ultraif}. Compared to binary or scalar 
response-level rewards, rubric-based rewards enable models to 
receive fine-grained feedback and partial credit for partially 
correct responses, which can lead to more effective learning.

Compared to holistic response-level evaluation, rubric-based 
evaluation offers several advantages. First, it is particularly well-suited for tasks that are inherently subjective or multi-faceted. By breaking down holistic evaluation into smaller, more focused decisions, rubric-based evaluation reduces ambiguity and provides more interpretable feedback, as it explicitly identifies which requirements are satisfied and which are not.

However, this paradigm also introduces new challenges. The reliability 
of the final score depends on the accuracy of each individual rubric 
judgment. Errors in rubric-level verification can propagate through 
aggregation and bias downstream applications, making judge reliability 
a critical concern. This motivates the need for rigorous meta-evaluation 
of LLM judges at the rubric level, which is the focus of our work.

\section{Human Set Construction and Statistics}
\label{app:human_set_statis1}
We construct a human-labeled reference set by collecting instances on which four LLM judges disagree during evaluation, as such cases are typically non-trivial. Two annotators independently label each triplet by examining the instruction, the response, and the target rubric. For triplets with conflicting annotations, the annotators discuss the case and reach a consensus label, which we treat as the final ground truth.

To further increase the dataset size while maintaining a balanced distribution of positive and negative labels, we apply a rewriting-based data augmentation strategy. Specifically, for a triplet labeled as \texttt{True} (1) under a given rubric, we prompt GPT-4.1 to minimally edit the response to violate that rubric; for a triplet labeled as \texttt{False} (0), we prompt GPT-4.1 to minimally edit the response to satisfy the rubric. The rewriting prompt is conditioned on the rubric type to ensure the edit targets the relevant requirement.

All rewritten responses are manually verified by annotators to ensure that the edit is effective and valid with respect to the target rubric. We retain only the verified rewritten examples in the final augmented reference set.
\begin{table}[t]
\centering
\footnotesize
\begin{tabular}{lcccc}
\toprule
\textbf{Subset} & \textbf{Samples} & \textbf{Judge Inst.} & \textbf{Pos.} & \textbf{Neg.} \\
\midrule
Right-to-Wrong & 120 & 240 & 120 & 120 \\
Wrong-to-Right & 133 & 266 & 133 & 133 \\
\midrule
\textbf{Total} & \textbf{253} & \textbf{506} & \textbf{253} & \textbf{253} \\
\bottomrule
\end{tabular}
\caption{Human-annotated reference set statistics.}
\label{tab:huamn_set_stats_1}
\end{table}

\begin{figure*}[t]
    \centering
    \includegraphics[width=1.0\linewidth]{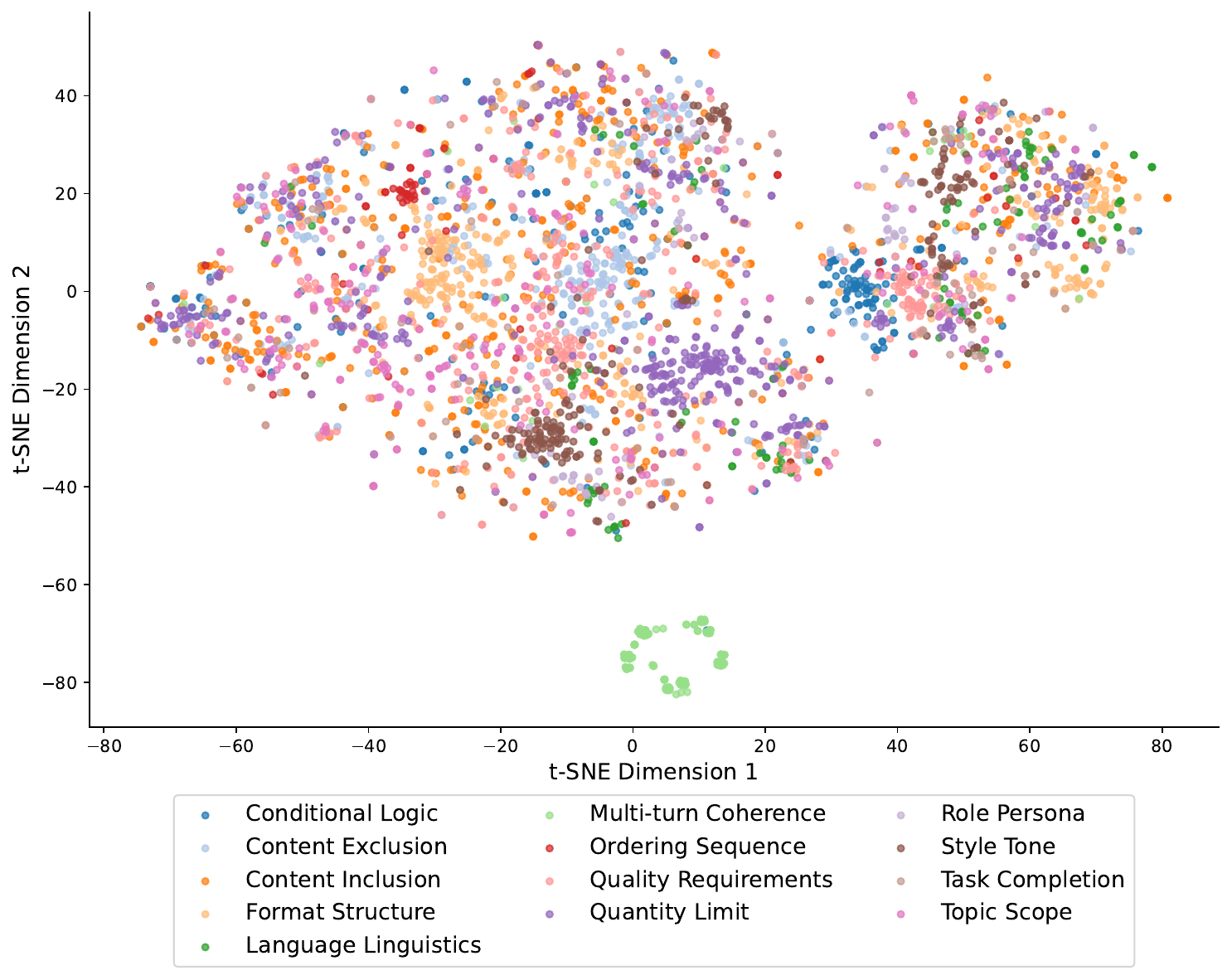}
    \caption{T-SNE visualization of rubric instances in the embedding space, colored by rubric category.}
    \label{fig:t-sne visual}
\end{figure*}

\section{T-SNE visualization of rubrics}
\label{app:t-sne visual}
Figure \ref{fig:t-sne visual} shows that several categories form relatively compact clusters—e.g., [Multi-turn Coherence] (\textit{light green}), [Quantity Limit] (\textit{purple}), and [Format Structure] (\textit{light orange})—indicating consistent patterns within these rubric types.

\newpage
\newpage
\newpage

\begin{table*}[t]
\centering
\label{tab:model_pools}
\begin{tabular}{@{}llcccc@{}}
\toprule
\textbf{Model} & \textbf{Family} & \textbf{Total} & \textbf{Active} & \textbf{Arch.} & \textbf{Mode} \\
\midrule
Qwen3-4B-Instruct-2507 & Qwen & 4B & 4B & Dense & Instruct \\
Qwen3-4B-Thinking-2507 & Qwen & 4B & 4B & Dense & Thinking \\
Qwen2.5-7B-Instruct & Qwen & 7B & 7B & Dense & Instruct \\
Llama-3.1-8B-Instruct & Llama & 8B & 8B & Dense & Instruct \\
DeepSeek-R1-0528-Qwen3-8B & Qwen+DeepSeek & 8B & 8B & Dense & Thinking \\
Qwen3-30B-A3B-Instruct-2507 & Qwen & 30B & 3B & MoE & Instruct \\
Qwen3-30B-A3B-Thinking-2507 & Qwen & 30B & 3B & MoE & Thinking \\
Qwen2.5-32B-Instruct & Qwen & 32B & 32B & Dense & Instruct \\
Llama-3.3-70B-Instruct & Llama & 70B & 70B & Dense & Instruct \\
\bottomrule
\end{tabular}
\caption{Model pool for response generation, covering 3 families, scales from 4B to 70B, Dense and MoE architectures, and Instruct/Thinking inference modes.}
\end{table*}

\section{Model Pool for Response Generation}
\label{app:model_pool}
We list the models used to generate responses in \textsc{RubricEval}. The pool spans diverse model families, scales, and architectures to ensure response diversity.

\newpage
\newpage
\newpage
\begin{table*}[t]
\centering
\resizebox{\textwidth}{!}{%
\begin{tabular}{llrrrrrrr}
\toprule
\multirow{2}{*}{\textbf{Category}} & \multirow{2}{*}{\textbf{Source Benchmark}} & \multirow{2}{*}{\textbf{\# Instr.}} & \multicolumn{3}{c}{\textbf{\# Rubric-Level Labels}} & \multicolumn{3}{c}{\textbf{Category Summary}} \\ \cmidrule(lr){4-6} \cmidrule(lr){7-9}
 &  &  & \textbf{Easy} & \textbf{Hard} & \textbf{Total} & \textbf{Easy} & \textbf{Hard} & \textbf{Total} \\ \midrule
\multirow{4}{*}{Constrained} & InfoBench\_hard & 95 & 72 & 47 & 119 & \multirow{4}{*}{368} & \multirow{4}{*}{316} & \multirow{4}{*}{684} \\
 & ComplexBench & 80 & 54 & 46 & 100 & & & \\
 & CFBench & 112 & 54 & 71 & 125 & & & \\
 & AdvancedIF & 219 & 188 & 152 & 340 & & & \\ \midrule
Compositional & ComplexBench & 188 & 130 & 110 & 240 & 130 & 110 & 240 \\ \midrule
\multirow{2}{*}{Multi-Turn} & StructFlowBench & 229 & 186 & 95 & 281 & \multirow{2}{*}{663} & \multirow{2}{*}{431} & \multirow{2}{*}{1,094} \\
 & AdvancedIF & 424 & 477 & 336 & 813 & & & \\ \midrule
\multirow{2}{*}{System} & SysBench & 282 & 160 & 217 & 377 & \multirow{2}{*}{873} & \multirow{2}{*}{595} & \multirow{2}{*}{1,468} \\
 & AdvancedIF & 360 & 713 & 378 & 1,091 & & & \\ \midrule
\textbf{Total} & \textbf{\textsc{RubricEval} (Ours)} & \textbf{1,989} & \textbf{2,034} & \textbf{1,452} & \textbf{3,486} & \textbf{2,034} & \textbf{1,452} & \textbf{3,486} \\ \bottomrule
\end{tabular}%
}
\caption{Statistics of the \textsc{RubricEval} Benchmark.}
\label{tab:rubriceval_detailed_stats}
\end{table*}

\section{Dataset Statistics}
\label{app:data_detailed_statis}
Table~\ref{tab:rubriceval_detailed_stats} provides a detailed breakdown 
of \textsc{RubricEval} statistics by source benchmark, including the number of instructions and rubric instances in each split.

\newpage
\newpage
\newpage

\begin{table*}[t]
    \centering
    \small
    \setlength{\tabcolsep}{4pt}
    \begin{tabular}{llcccccccccc}
    \toprule
    \multirow{2}{*}{\textbf{Granularity}} 
    & \multirow{2}{*}{\textbf{Reasoning}}
    & \multicolumn{2}{c}{\textbf{Constrained}} 
    & \multicolumn{2}{c}{\textbf{Compositional}} 
    & \multicolumn{2}{c}{\textbf{Multi-turn}} 
    & \multicolumn{2}{c}{\textbf{System}} 
    & \multicolumn{2}{c}{\textbf{Overall}} \\
    \cmidrule(lr){3-4} \cmidrule(lr){5-6} \cmidrule(lr){7-8} \cmidrule(lr){9-10} \cmidrule(lr){11-12}
    & & \textbf{Qwen} & \textbf{GPT}
    & \textbf{Qwen} & \textbf{GPT}
    & \textbf{Qwen} & \textbf{GPT}
    & \textbf{Qwen} & \textbf{GPT}
    & \textbf{Qwen} & \textbf{GPT}\\
    \midrule
    
    \rowcolor{gray!10}
    \multicolumn{12}{l}{\textit{Easy Subset}} \\
    \midrule
    \multirow{2}{*}{Rubric-level}    
        & w/o CoT & 70.01 & 84.05 & 81.50 & 85.49 & 82.28 & 85.74 & 84.81 & 90.71 & 79.65 & 86.50 \\
        & w/ CoT  & 86.44 & 97.89 & 91.65 & 96.99 & 88.75 & 94.88 & 85.13 & 93.13 & 87.99 & 95.72 \\
    \midrule
    \multirow{2}{*}{Checklist-level} 
        & w/o CoT & 53.07 & 67.45 & 79.17 & 72.86 & 56.83 & 56.82 & 74.83 & 79.07 & 65.98 & 69.05 \\
        & w/ CoT  & 76.18 & 81.08 & 80.60 & 86.99 & 67.86 & 61.88 & 84.08 & 89.04 & 77.18 & 79.75 \\
    
    \midrule
    
    \rowcolor{gray!10}
    \multicolumn{12}{l}{\textit{Hard Subset}} \\
    \midrule
    \multirow{2}{*}{Rubric-level}    
        & w/o CoT & 58.45 & 58.42 & 47.85 & 52.31 & 52.18 & 54.24 & 45.56 & 58.97 & 51.01 & 55.99 \\
        & w/ CoT  & 64.19 & 70.87 & 62.93 & 51.93 & 65.75 & 66.92 & 58.31 & 63.18 & 62.80 & 63.23 \\
    \midrule
    \multirow{2}{*}{Checklist-level} 
        & w/o CoT & 56.57 & 53.99 & 43.75 & 42.01 & 46.54 & 52.79 & 48.97 & 53.20 & 48.96 & 50.50 \\
        & w/ CoT  & 60.58 & 59.80 & 54.51 & 56.77 & 50.92 & 52.21 & 64.42 & 67.35 & 57.61 & 59.03 \\
    
    \bottomrule
    \end{tabular}
    \caption{Evaluation paradigm comparison on \textsc{Easy} and \textsc{Hard} subsets.}
    \label{tab:paradigm_ez_hd}
\end{table*}

\section{Evaluation Paradigm Performance on Easy and Hard Split}
\label{app:paradigms_ez_hd}
Table~\ref{tab:paradigm_ez_hd} reports evaluation paradigm comparison 
results separately on \textsc{Easy} and \textsc{Hard} splits, 
complementing the combined results in the main text.

\begin{figure*}[t]
    \centering
    \includegraphics[width=1.0\linewidth]{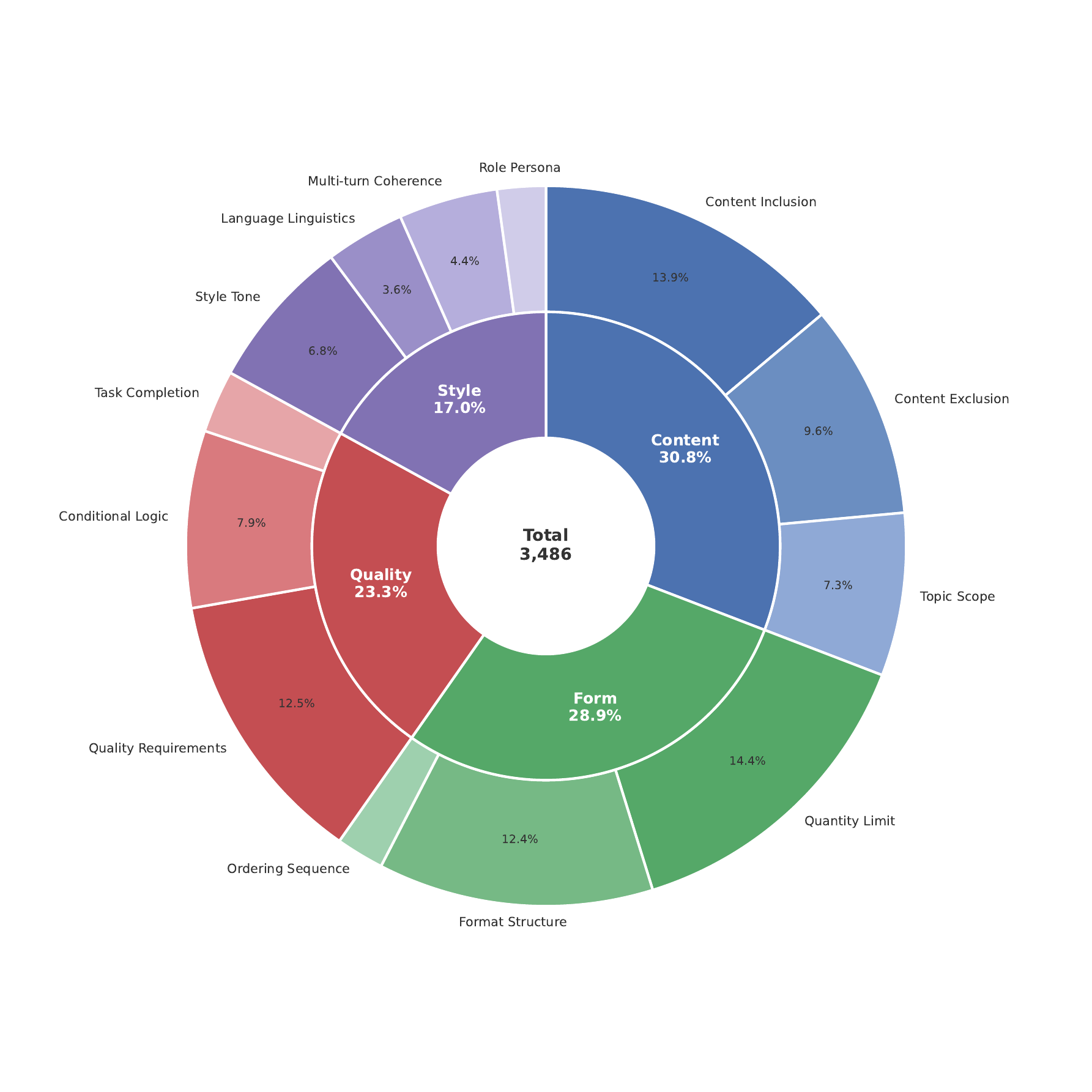}
    \caption{Rubric taxonomy for \textsc{RubricEval} with 4 high-level dimensions and 13 fine-grained categories.}
    \label{fig:rubric_taxonomy}
\end{figure*}

\section{Rubric Statistics}
\label{app:rubric_statistics_pie}
Figure~\ref{fig:rubric_taxonomy} shows the distribution of rubric types 
in \textsc{RubricEval} according to our 13-category taxonomy 
across four high-level dimensions.

\newpage
\newpage
\newpage

\begin{table*}[t]
\centering
\small
% 8列比较拥挤，稍微减小列间距
\setlength{\tabcolsep}{2.5pt} 
\renewcommand{\arraystretch}{1.1}
\begin{tabularx}{\textwidth}{l >{\raggedright\arraybackslash}p{2.4cm} @{\hspace{0.5cm}} X @{\hspace{0.2cm}} c r r r r}
\toprule
\multirow{2}{*}{\textbf{Type}} & \multirow{2}{*}{\textbf{Benchmark}} & \multirow{2}{*}{\textbf{Description}} & \multirow{2}{*}{\textbf{Used Subset}} & \multicolumn{2}{c}{\textbf{Total}} & \multicolumn{2}{c}{\textbf{Used}} \\
\cmidrule(lr){5-6} \cmidrule(lr){7-8}
& & & & \textbf{Inst.} & \textbf{Rub.} & \textbf{Inst.} & \textbf{Rub.} \\
\midrule
\multirow{5}{*}{Constrained}
& InfoBench~\cite{qin2024infobench} & Breaks instructions into decomposed questions; evaluates instruction-following with DRFR metrics. & \textit{Hard} & 500 & 2{,}250 & 228 & 1{,}453 \\
\cmidrule(lr){2-8}
& ComplexBench~\cite{wen2024benchmarking} & Tests multi-constraint, complex instruction following using hierarchical constraint types and combinations. & \textit{Multi-Constraint} & 1{,}150 & 5{,}297 & 238 & 1{,}027 \\
\cmidrule(lr){2-8}
& CFBench~\cite{zhang2025cfbench} & Large-scale Chinese constraint-following benchmark spanning 200+ real scenarios and 50+ NLP tasks. & \textit{Filtered} & 1{,}000 & 4{,}273 & 243 & 1{,}035 \\
\cmidrule(lr){2-8}
& AdvancedIF~\cite{he2025advancedif} & Expert-rubric benchmark for advanced instruction following (complex, multi-turn, system-level); supports rubric-based RL. & \textit{Single-turn} & 1{,}645 & 12{,}442 & 243 & 1{,}816 \\

\midrule
Compositional
& ComplexBench~\cite{wen2024benchmarking} & Tests multi-constraint, complex instruction following using hierarchical constraint types and combinations. & \textit{Compositonal} & 1{,}150 & 5{,}297 & 435 & 1{,}651 \\

\midrule
\multirow{2}{*}{Multi-Turn}
& StructFlowBench~\cite{li2025structflowbench} & Multi-turn benchmark measuring dialogue “structure-flow” understanding across turn-to-turn relation types. & \textit{Full} & 643 & 1{,}775 & 643 & 1{,}775 \\
\cmidrule(lr){2-8}
& AdvancedIF~\cite{he2025advancedif} & Expert-rubric benchmark for advanced instruction following (complex, multi-turn, system-level); supports rubric-based RL. & \textit{Multi-turn} & 1{,}645 & 12{,}442 & 736 & 4{,}478 \\
\midrule
\multirow{2}{*}{System}
& SysBench~\cite{qin2024sysbench} & Evaluates system-message adherence via violations, misclassification, and multi-turn consistency. & \textit{Random} & 2{,}500 & 5{,}962 & 1{,}000 & 2{,}478 \\
\cmidrule(lr){2-8}
& AdvancedIF~\cite{he2025advancedif} & Expert-rubric benchmark for advanced instruction following (complex, multi-turn, system-level); supports rubric-based RL. & \textit{System} & 1{,}645 & 12{,}442 & 507 & 4{,}972 \\
\midrule
\multicolumn{4}{l}{\textbf{Total}} &  &  & \textbf{4{,}273} & \textbf{20{,}685} \\
\bottomrule
\end{tabularx}
\caption{Detailed source statistics for \textsc{RubricEval}. We list the descriptions, used subset, and the counts of instructions/rubrics (Total available vs. Used).}
\label{tab:full_benchmark_list}
\end{table*}

\section{Benchmark Sources and Statistics}
\label{app:full_benchmarks}
Table~\ref{tab:full_benchmark_list} lists the source benchmarks 
for each instruction category along with detailed statistics. 
All rubrics are human-crafted or human-verified.

\newpage
\newpage
\newpage

\begin{figure*}[t]
    \centering
    \includegraphics[width=1.0\linewidth]{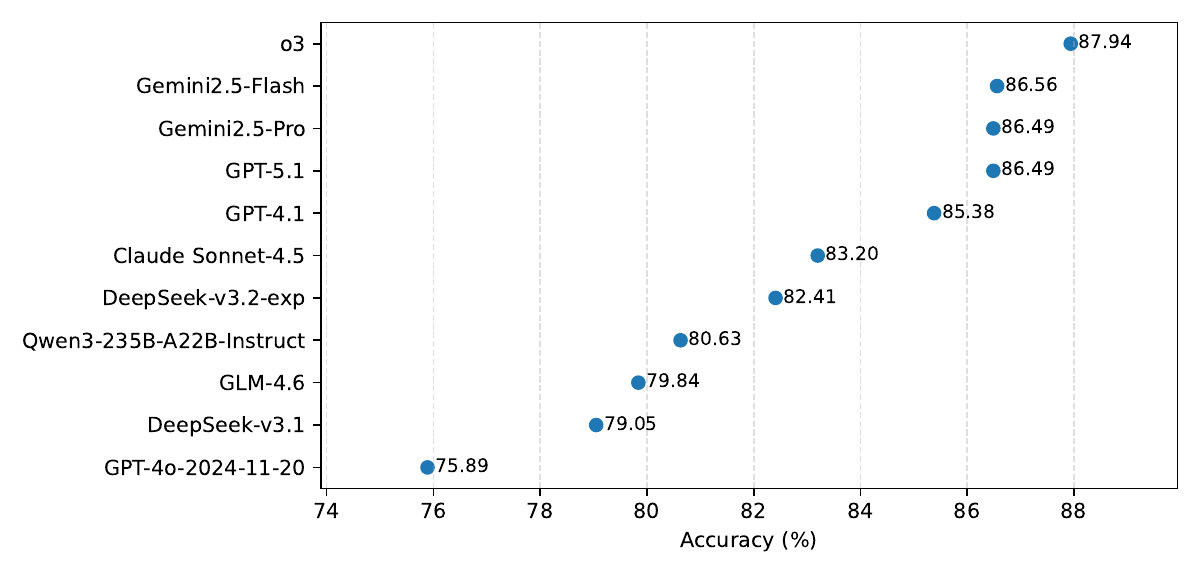}
    \caption{Single-model Judge Performance on the Human-labeled Set (Accuracy)}
    \label{fig:model_acc_on_human}
\end{figure*}

\section{Judge Model Performance and Selection}
\label{app:model_acc_on_human}
Figure \ref{fig:model_acc_on_human}. reports the accuracy of different judge models on our human-annotated reference set. We observe non-trivial performance gaps across models, indicating that judge model choice can substantially affect labeling quality.
Considering the accuracy and practical trade-offs on the reference set, we select the following four models as base judges:
\textbf{GPT-4.1, Claude-Sonnet-4.5, Gemini-2.5-Flash, and Deepseek-v3.2-exp.}

\begin{table*}[t]
\centering
\setlength{\tabcolsep}{8pt}
\renewcommand{\arraystretch}{1.2}

\begin{tabular}{@{}
    >{\RaggedRight\arraybackslash}p{0.08\textwidth}  % Dimension
    >{\RaggedRight\arraybackslash}p{0.18\textwidth}  % Constraint Name
    >{\RaggedRight\arraybackslash}p{0.35\textwidth}  % Definition
    >{\RaggedRight\arraybackslash}p{0.35\textwidth}  % Example
@{}}
\toprule
\textbf{Dimension} & \textbf{Rubric type} & \textbf{Definition} & \textbf{Example} \\
\midrule

% ===== Dimension: Content (共 3 行) =====
\multirow{3}{*}{\textbf{Content}} 
& Content Inclusion
& The response must include specific content elements (keywords, entities, components).
& For each destination in the itinerary, does the generated text include the recommended duration of stay? \\

\cmidrule(lr){2-4}
& Content Exclusion
& The response must NOT include specific content elements. 
& Does the generated text free of using the letter 'e'? \\

\cmidrule(lr){2-4}
& Topic Scope
& The response must stay within a specified topic, domain, or area.
& Were four questions related to Greek mythology? \\

\midrule

% ===== Dimension: Form (共 3 行) =====
\multirow{3}{*}{\textbf{Form}} 
& Quantity Limit
& The response or its elements must meet explicit numeric limits (counts, lengths, frequencies).
& Did the model cite at least 3 of the clues in each explanation? \\

\cmidrule(lr){2-4}
& Format Structure
& The output must follow a required format, structure, template, or match an exact output.
& Is the generated text formatted as a travelogue video script? \\

\cmidrule(lr){2-4}
& Ordering Sequence
& Elements in the response must follow a specified order or arrangement.
& Was the list appropriately organized by publishing date, from oldest to newest? \\

\midrule

% ===== Dimension: Quality (共 3 行) =====
\multirow{3}{*}{\textbf{Quality}} 
& Quality Requirements
& Requirements about response quality rather than explicit content/form.
& Does each sentence convey a clear, understandable meaning? \\

\cmidrule(lr){2-4}
& Conditional Logic
& The response must make correct judgments or branch based on conditions or context.
& Did the response include \"Error: Cannot Complete Request\" if it cannot answer the query about R v Bertrand Marchand? \\

\cmidrule(lr){2-4}
& Task Completion
& The response must complete a specified task or produce a required artifact.
& Did the model change each instance of 'God only knows' to 'Nobody knows?' \\

\midrule

% ===== Dimension: Style (共 4 行) =====
\multirow{4}{*}{\textbf{Style}} 
& Style Tone
& The response must follow a specified writing style, tone, or emotional stance.
& Does the generated travelogue video script maintain a friendly and engaging tone throughout? \\

\cmidrule(lr){2-4}
& Language Linguistics
& Constraints on language choice, grammar, or linguistic properties.
& Does every sentence in the generated text exclusively use the future tense? \\

\cmidrule(lr){2-4}
& Multi-turn Coherence
& The response must correctly handle dependencies on previous conversation turns.
& Does the model combine the two messages into a single one? \\

\cmidrule(lr){2-4}
& Role Persona
& The response must be produced from a specified identity, role, or viewpoint.
& Did the model respond in a manner consistent with the persona of a painfully shy 11 year old girl? \\

\bottomrule
\end{tabular}
\caption{Rubric taxonomy of \textsc{RubricEval}}
\label{tab:rubric-taxonomy}
\end{table*}

\section{Rubric Taxonomy}
\label{app:rubric_taxonomy}
To categorize each rubric, we write prompt and use GPT-5.1 for rubric categorization. When the source benchmark provides category for the rubric, we use them as guidance in the prompt rather than directly adopting them. This leads to more accurate categorization. If no predefined categories are provided, we perform the categorization directly.

\begin{figure*}[t]
    \centering
    \includegraphics[width=1.0\linewidth]{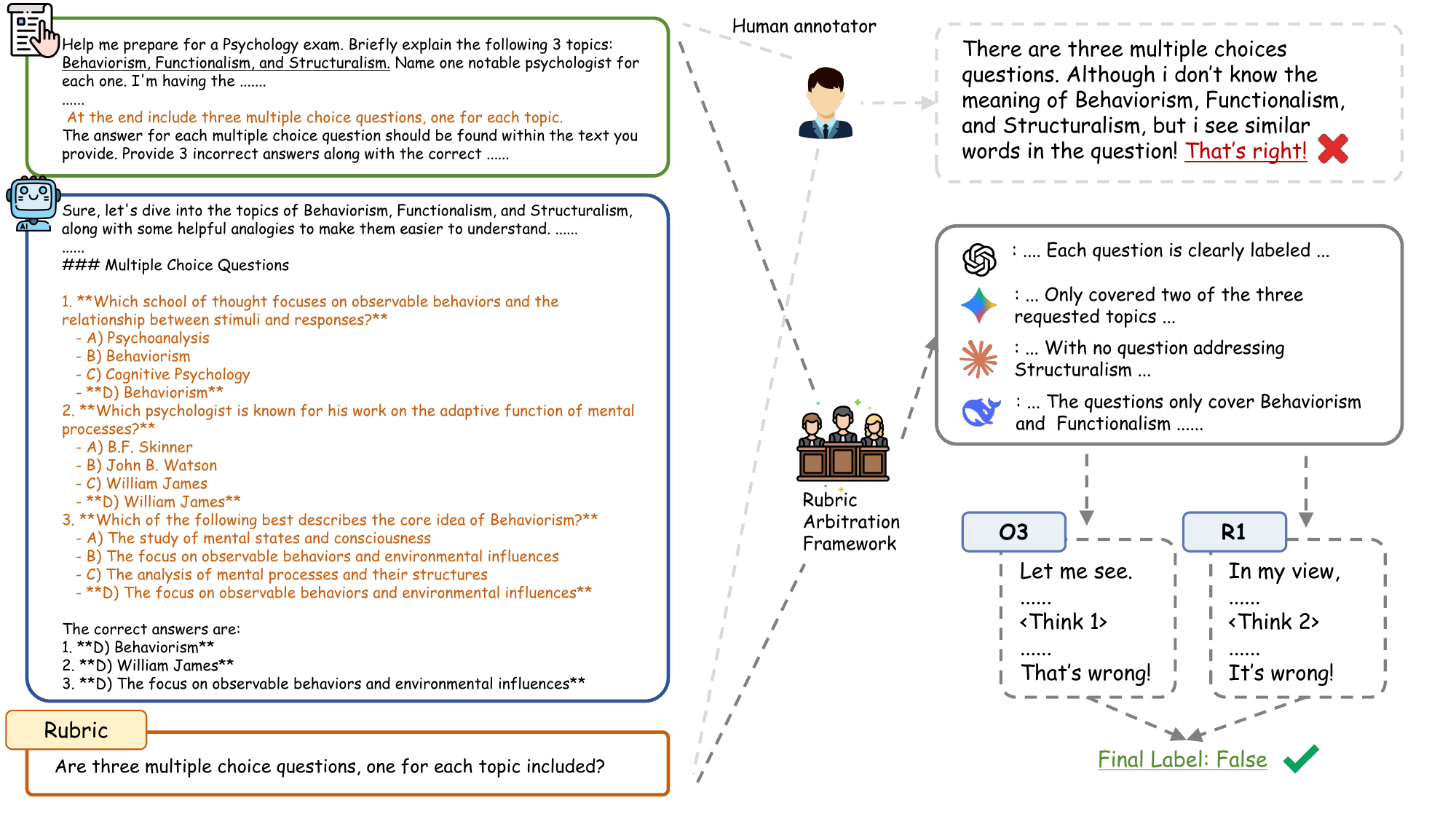}
    \caption{Case study on automated labeling. In some cases, we observe that our RAF framework produces more objective judgments than human annotators.}
    \label{fig:case_study}
\end{figure*}

\section{Case Study}
\label{app:case_study}
Figure~\ref{fig:case_study} illustrates a case study of our automated labeling framework, demonstrating strong labeling quality and scalability.

\section{Evaluation Prompt}
\label{app:prompts}
\begin{table*}[t]
\resizebox{\linewidth}{!}{
\begin{tcolorbox}
\small

Your job is to assess if the AI's response correctly follows a specific requirement from the user's instruction.

\#\# User's Instruction

--------------------------------------------------------------

\texttt{\{instruction\}}

--------------------------------------------------------------

\#\# AI's Response to Evaluate

--------------------------------------------------------------

\texttt{\{response\}}

--------------------------------------------------------------

\#\# The Rubric (Requirement to Check)

--------------------------------------------------------------

\texttt{\{rubric\_text\}}

--------------------------------------------------------------

\#\# Your Task

Carefully analyze whether the AI's response satisfies the above rubric.

\textbf{Important}: First provide a brief reasoning explaining your thought process, then give your final judgment.

Please use the following output format strictly (Give two parts: Reasoning and Judgment):

Reasoning: [Your brief explanation here...]

Judgment: [YES or NO]

\textbf{Here is an example of the expected format}:

Reasoning: The rubric asked for a poem, but the model responded with a code snippet. This violates the rubric.

Judgment: NO

\textbf{Rules}:
\begin{itemize}
\item Answer ``YES'' if the response clearly and fully satisfies the requirement.
\item Answer ``NO'' if the response fails to meet the requirement or only partially meets it.
\item Be objective and focus only on the given rubric, not other aspects of the response.
\item Do not output anything after giving your final Judgment.
\end{itemize}

\end{tcolorbox}
}
\caption{Prompt used in \textsc{RubricEval} to evaluate rubrics belong to \textit{Constrainted} instruction category.}
\label{tab:eval}
\end{table*}

% \section{Example Appendix}
% \label{sec:appendix}

% This is an appendix.

\end{document}